\documentclass{article}
\usepackage{arxiv}
\usepackage[utf8]{inputenc} 
\usepackage[T1]{fontenc}    
\usepackage[pdfborder={0 0 0}]{hyperref}       
\usepackage{url}            
\usepackage{booktabs}       
\usepackage{amsfonts}       
\usepackage{nicefrac}       
\usepackage{microtype}      
\usepackage{lipsum}     
\usepackage{graphicx}
\usepackage{natbib}
\usepackage{doi}
\usepackage[most]{tcolorbox}
\title{Open-Source LLMs for Text Annotation: A Practical Guide for Model Setting and Fine-Tuning}

\author{
{Meysam Alizadeh}\thanks{\normalsize{Corresponding author (alizadeh@ipz.uzh.ch)}.}\\
University of Zurich\\
Zurich, Switzerland\\
\And
{Maël Kubli}\\
University of Zurich\\
Zurich, Switzerland\\
\And
{Zeynab Samei}\\
Institute for Fundamental Research\\
Tehran, Iran\\
\And
{Shirin Dehghani}\\
Allameh Tabataba'i University\\
Tehran, Iran\\
\And
{Mohammadmasiha Zahedivafa}\\
Iran University of Science and Technology\\
Tehran, Iran \\
\And
{Juan D. Bermeo}\\
University of Zurich\\
Zurich, Switzerland\\
\And
{Maria Korobeynikova}\\
University of Zurich\\
Zurich, Switzerland\\
\And
{Fabrizio Gilardi}\\
University of Zurich\\
Zurich, Switzerland\\
}

\renewcommand{\headeright}{}
\renewcommand{\undertitle}{}
\renewcommand{\shorttitle}{Open-Source LLMs for Text Annotation}

\hypersetup{
pdftitle={Open-Source LLMs for Text Annotation: A Practical Guide for Model Setting and Fine-Tuning},
pdfsubject={},
pdfauthor={Meysam Alizadeh, Maël Kubli, Zeynab Samei, Shirin Dehghani, Mohammadmasiha Zahedivafa, Juan D. Bermeo, Maria Korobeynikova, Fabrizio Gilardi},
pdfkeywords={ChatGPT, LLMs, Large Language Models, NLP, Text Annotation, Open Source, Fine Tuning},
}

\begin{document}

\maketitle

\begin{abstract}
This paper studies the performance of open-source Large Language Models (LLMs) in text classification tasks typical for political science research. By examining tasks like stance, topic, and relevance classification, we aim to guide scholars in making informed decisions about their use of LLMs for text analysis. Specifically, we conduct an assessment of both zero-shot and fine-tuned LLMs across a range of text annotation tasks using news articles and tweets datasets. Our analysis shows that fine-tuning improves the performance of open-source LLMs, allowing them to match or even surpass zero-shot GPT-3.5 and GPT-4, though still lagging behind fine-tuned GPT-3.5. We further establish that fine-tuning is preferable to few-shot training with a relatively modest quantity of annotated text. Our findings show that fine-tuned open-source LLMs can be effectively deployed in a broad spectrum of text annotation applications. We provide a Python notebook facilitating the application of LLMs in text annotation for other researchers.
\end{abstract}

\newpage
\section{Introduction}
Generative Large Language Models (LLMs) such as GPT-3 and GPT-4 have demonstrated substantial potential for text-annotation tasks common to many Natural Language Processing (NLP) and political science applications \citep{Ding:2023aa}. Recent research reports impressive performance metrics for these models. For instance, studies demonstrate that GPT-3.5 exceeds the performance of crowd-workers in tasks encompassing relevance, stance, sentiment, topic identification, and frame detection \citep{gilardi2023chatgpt}, that it outperforms trained annotators in detecting the political party affiliations of Twitter users \citep{Tornberg:2023aa}, and that it achieves accuracy scores over 0.6 for tasks such as stance, sentiment, hate speech detection, and bot identification \citep{Zhu:2023aa}. Notably, GPT-3.5 also demonstrates the ability to correctly classify more than 70\% of news as either true or false \citep{Hoes:2023aa}, which suggests that LLMs might potentially be used to assist content moderation processes.

While the performance of LLMs for text annotation is promising, several aspects remain unclear and require further research. Before proceeding to classify documents, the researcher must: (1) choose whether directly use a LLM without any additional training (zero-shot) or to manually annotate some data and follow a few-shot or fine-tuning approach, (2) decide how many instances of data to annotate for few-shot or fine-tuning, (3) choose between GPT-3.5, GPT-4, and open-source LLMs such as LLaMA and FLAN. In the sections below, we offer empirical evidence demonstrating the significance of these decisions in practical terms and provide recommendations on how to best make their decisions. Throughout, our goal is to highlight the capabilities of open-source LLMs and fine-tuning approach, and provide a practical guide for researchers.

Zero-Shot Learning enables models to generalize to hitherto unseen tasks without the requirement for labeled examples, while Few-Shot Learning leverages a minimal set of annotated instances to adapt the model to new tasks. Despite their applicability, the conditions under which one paradigm outperforms the other remain an open question. 
On the other hand, fine-tuning constitutes the retraining of LLMs on a specialized, domain-specific dataset to augment task-specific performance. This process helps in endowing the LLM with domain-specific knowledge, while also potentially reducing some of the model biases. However, this may introduce biases present in the fine-tuning dataset, thereby requiring careful consideration of the data employed for this purpose. 
The extent to which Fine-Tuning and Few-Shot Learning methodologies substantively improve model performance remains indeterminate. Specifically, unresolved issues include the quantity of annotated data requisite for significant performance gains in the Fine-Tuning process, and whether Few-Shot Learning yields statistically meaningful improvements.

Moreover, the role of open-source LLMs deserves more attention. While models like GPT-3.5 have democratized the field by offering a more cost-effective alternative to traditionally more expensive annotation methods involving human annotations, open-source LLMs represent a further step towards greater accessibility. Beyond cost, the advantages of open-source LLMs include degrees of transparency and reproducibility that are typically not provided by commercial models. open-source LLMs can be scrutinized, tailored, and enhanced by a wider user base, fostering a diverse group of contributors and improving the overall quality and fairness of the models. Furthermore, open-source LLMs offer significant data protection benefits. They are designed not to share data with third parties, enhancing security and confidentiality. For these reasons, the academic community is increasingly advocating for the use of open-source LLMs \citep{Spirling:2023aa,Ollion:2024aa}. This transition would not only broaden researchers' access to these tools but also promote a more open and reproducible research culture.

To address these questions, we extend previous research \citep{gilardi2023chatgpt} to compare the performance of two widely-used open-source LLMs, LLaMA and FLAN, with that of GPT 3.5 as well as MTurkers, using eleven text annotation tasks distributed across four datasets. Each model is tested using different settings: varied model sizes for FLAN, and distinct temperature parameters in both zero-shot, few-shot, and fine-tuning approaches for GPT 3.5, LLaMA-1 (through HuggingChat), and LLaMA-2. We then compare their accuracy, using agreement with trained annotators as a metric, against that of MTurk as well as amongst themselves. 


\section{Related Work}
The practice of fine-tuning transformer models for specialized tasks has become a cornerstone in the field of natural language processing (NLP). With the advent of large language models (LLMs) like GPT-3.5/4 and Bard (now Gemini), a burgeoning body of research is emerging to evaluate their performance and utility across various tasks.

LLMs have catalyzed a significant paradigmatic shift in the field of NLP, notably in the area of text annotation. Their capability to mimic human behavior, understand context, and adapt makes them effective for data annotation tasks \citep{yang2023harnessing}. This adaptability is primarily conditioned through intricate prompt engineering methods, ranging from zero-shot to few-shot promoting techniques, thereby focusing on generating accurate outputs \citep{liu2023pre}. 

Previous studies have extensively explored the capabilities of LLMs in diverse NLP applications such as text classification, text alignment, and semantic similarity tasks. \citep{gilardi2023chatgpt} has shown that GPT-3.5 excels in multiple annotation tasks compared to human annotators. \citep{tornberg2023chatgpt} further substantiates the efficacy of LLMs in the political arena by demonstrating that ChatGPT-4 outperforms both expert classifiers and crowd workers in classifying the political affiliation of Twitter posts, even when requiring reasoning based on contextual knowledge and author intentions. Promising results in named entity recognition, fact-checking and various text annotation tasks have also been reported \citep{frei2023annotated, Ding:2023aa, hoes2023using}. However, the focus has largely been on practical NLP applications, leaving unexplored the potential in corpus pragmatics and corpus-assisted discourse studies \citep{yang2023harnessing}.

However, it is imperative to acknowledge that the performance of LLMs is highly contingent upon both the quality of the dataset provided to be annotated and the proficiency of the model itself in this area \citep{pangakis2023automated}. Therefore, the ongoing refinement and fine-tuning of these models are imperative.

In terms of fine-tuning methodologies, various approaches have been developed. Notably, Parameter-Efficient Fine Tuning (PEFT) has garnered attention for optimizing LLMs in resource-constrained settings \citep{hu2023llm}. Another method that has received scholarly attention is the 'chain of thought' technique, as validated by \citep{zhang2023multimodal, wei2022chain}. The prevailing methodology, however, continues to be the attachment of task-specific heads to existing architectures, followed by domain-specific training and performance evaluation \citep{howard2018universal}. Moreover, several papers suggest that fine-tuning large language models can be effective in improving their performance and also reduce their size.  \citet{wang-etal-2020-structured} propose a structured pruning approach based on low-rank factorization and L0 norm regularization, which achieves significant inference speedups while maintaining or surpassing the performance of unstructured pruning methods. \citet{binz2023turning} explores the possibility of turning large language models into cognitive models by fine-tuning them on psychological experiment data, showing that they can accurately represent human behavior and outperform traditional cognitive models in decision-making tasks.

The ascendancy of proprietary LLMs has engendered a series of ethical and practical concerns, particularly pertaining to cost, transparency, and data protection. In contrast, open-source LLMs offer compelling advantages, including cost-effectiveness, methodological transparency, replicability, and stringent data protection standards \citep{liesenfeld2023opening, Spirling:2023aa,Ollion:2024aa}.

Given the rapidly evolving landscape of LLM applications in text annotation and the critical role of prompt engineering in their performance, our work aims to evaluate the efficiency of fine-tuned open-source LLMs across a broader range of text annotation tasks. We also offer a side-by-side performance comparison with fine-tuned GPT-3.5.

\section{Materials and Methods} 

\subsection{Data}

The analysis relies on four distinct datasets. The first dataset consists of 2,978 randomly selected tweets from a more extensive collection of 2.6 million tweets related to content moderation, spanning from January 2020 to April 2021. The second dataset comprises 3,006 tweets posted by members of the US Congress between 2017 and 2022, sampled from a dataset of 20 million tweets. The third dataset consists of 2,480 newspaper articles on content moderation published from January 2020 to April 2021, drawn from a dataset of 980k articles obtained via LexisNexis. Sample sizes were determined based on the number of texts required to construct training sets for machine-learning classifiers. Finally, the fourth dataset replicates the data collection process of the first dataset. Specifically, it focused on January 2023, comprising a random sample of 1,313 Tweets from a dataset of 1.3 million tweets.

For the fine-tuning section of the three LLMs, we aim to allocate at least 15 \% or more of the dataset to the evaluation set and the remaining 85 \% to the training sets. We structure the training sets in increments of 50, 100, 250, 500, 1000, and 1500 samples, depending on the dataset's size. If the evaluation contains fewer than 100 rows or less than two instances of the minority class (the least frequent class), we adjust its proportion upwards until these conditions are fulfilled. Beyond meeting these prerequisites, the evaluation set proportion is incrementally expanded by 5 \% as long as it does not compromise the planned training set sizes. This approach aims to optimize the evaluation sample size across an array of training set dimensions, thereby enabling a comprehensive assessment of how varying training data volumes impact the performance of all three LLMs. This approach also facilitates a comparative analysis using identical datasets for zero-shot learning, few-shot learning, and fine-tuning. 

\begin{table}
    \center
    \setlength\tabcolsep{2.5pt}
    \caption{Comprehensive Overview of Datasets Employed for Fine-Tuning Models Across Varied Tasks: Content Moderation Tweets from 2021 [A], Content Moderation Tweets from 2023 [B], Content Moderation News Articles from 2021 [C], and Tweets from the U.S. Congress spanning 2017 to 2021 [D]. Specifications encompass Task Categories, Evaluation Dataset Dimensions, and Varied Sample Sizes Utilized in Fine-Tuning Assessment.} 
    \label{table:datasets}
    \begin{tabular}{llrr}
      \toprule
      Dataset & Task & Eval. Size & Fine-Tuning Size \\ 
      \midrule
      A & Relevance & 387 & 50,100,250,500,1000,1500 \\ 
      A & Problem/Solution & 328 & 50,100,250,500 \\
      A & Policy Frames & 843 & 50,100,250,500 \\
      A & Stance Detection & 277 & 50,100,250,500,1000 \\
      A & Topics & 307 & 50,100,250 \\
      B & Relevance & 144 & 50,100,250,500,1000 \\
      B & Problem/Solution & 100 & 50,100,250,500 \\
      C & Relevance & 559 & 50,100,250,500,1000,1500 \\ 
      C & Problem Solution & 196 & 50,100,250,500,1000 \\
      D & Relevance & 836 & 50,100,250,500,1000,1500 \\ 
      D & Policy Frames & 341 & 50,100,250,500 \\
  \bottomrule
\end{tabular}
\end{table}

\subsection{Data Annotation Tasks}\label{section:annotation}
We implemented several annotation tasks: (1) \textit{relevance}: whether a tweet is about content moderation or, in a separate task, about politics; (2) \textit{topic detection}: whether a tweet is about a set of six pre-defined topics (i.e. Section 230, Trump Ban, Complaint, Platform Policies, Twitter Support, and others); (3) \textit{stance detection}: whether a tweet is in favor of, against, or neutral about repealing Section 230 (a piece of US legislation central to content moderation); (4) \textit{general frame detection}: whether a tweet contains a set of two opposing frames  (``problem' and ``solution''). The solution frame describes tweets framing content moderation as a solution to other issues (e.g., hate speech). The problem frame describes tweets framing content moderation as a problem on its own as well as to other issues (e.g., free speech); (5) \textit{policy frame detection}: whether a tweet contains a set of fourteen policy frames proposed in \citep{card2015media}. The full text of instructions for the five annotation tasks is presented in Appendix S1. We used the exact same wordings for LLMs and MTurk.

\subsection{Trained Annotators}\label{section:annotators}

We trained three political science students to conduct the annotation tasks. For each task, they were given the same set of instructions described above and detailed in Appendix \ref{appendix:codebook_zeroshot}. Importantly, to minimize inter-coder discrepancies and enhance the robustness of the annotation process, each of the students operated independently and systematically annotated the dataset task by task, adhering to a uniform codebook and shared procedural guidelines.

\subsection{Crowd-workers}\label{section:mturkers}

To maintain a consistent comparative framework, we engaged workers from Amazon's Mechanical Turk (MTurk) to execute the identical tasks administered to trained human annotators and Large Language Models (LLMs). These MTurk workers operated under the same instructional guidelines elaborated in Appendix \ref{appendix:codebook_zeroshot}. To ensure the quality and reliability of annotations, we imposed several restrictions on worker eligibility. Specifically, we limited task access to individuals designated as "MTurk Masters" by Amazon. Additionally, these workers were required to have a Human Intelligence Task (HIT) approval rate exceeding 90 \% and a minimum of 50 approved HITs. We further restricted their geographic location to the United States. To mitigate the risk of undue influence from individual workers on the annotations for a specific task, we instituted a cap, ensuring that no single worker could contribute annotations to more than 20 \% of the tweets allocated to a given task. Similar to our approach with trained human annotators, each tweet underwent annotation by two distinct MTurk workers to bolster the integrity and robustness of the collected data.

\subsection{LLM Selection and Settings}\label{section:llm}
In our endeavor to evaluate the annotation performance and cost efficiency of various large language models (LLMs), we selected four distinct LLMs. The first model chosen was OPENAI's GPT-4, GPT-3.5 (`gpt-3.5-turbo' version), a proprietary, closed-source LLM. Complementing this, we incorporated Meta's LLaMA-1 (`oasst-sft-6-LLaMA-30b' version) and the more recent LLaMA-2 in two configurations: `LLaMA-2 13b' and `LLaMA-2 70b. The selection was rounded off with FLAN-T5, a model we opted for due to its demonstrated promise in prior research \citep{chung2022scaling, ziems2023can}. For FLAN-T5, available in sizes ranging from 80M to 20B parameters, we experimented with the L, XL, and XXL variants to explore zero-shot capabilities (see Figure \ref{flan_comp} in Appendix). Ultimately, we selected the FLAN-XL model for fine-tuning due to its advantageous balance of computational resource demands and text processing capabilities. This selection was driven by the model's ability to provide a sophisticated understanding and processing of text, which is essential for optimal annotation performance.


In our Zero-Shot versus Few-Shot analysis, we employed only GPT-3.5 and LLaMA-1 (via HuggingChat) with the \textit{temperature} set to 0.2. For the fine-tuning phase, we utilized the bare LLaMA-1 model, FLAN-XL, and other selected models, with the \textit{temperature} set to 0.0. Our earlier findings informed this decision, where we observed a high level of agreement between runs with a \textit{temperature} of 0.2, eliminating the need to run each model twice. Adopting a \textit{temperature} of 0.0 for fine-tuning ensured the maximum level of output determinism, thereby enabling a more effective and efficient comparison between zero-shot and fine-tuned model performances.

\subsection{Prompt Engineering}

For zero-shot tests, we intentionally avoided adding any prompt engineering to ensure comparability between LLMs and MTurk crowd-workers. After testing several variations, we decided to feed tweets one by one to GPT-3.5 using the following prompt: ``Here's the tweet I picked, please label it as [Task Specific Instruction (e.g. `one of the topics in the instruction')].'' The corresponding prompts for each task are reported in Appendix \ref{appendix:codebook_fewshot_cot}. For few-shot tests, we employ Chain-of-Thought (CoT) prompting \citep{wei2022chain}, where large language models (LLMs) are provided with both the question and a step-by-step reasoning answer as examples. Specifically, following previous research \citep{kojima2022large}, we use GPT-3.5 to generate two CoT-prompted examples per class per annotation task. More specifically, we supplied GPT-3.5 with examples annotated by human experts, requesting an annotation and a substantiating explanation for the given annotation. Should the annotation provided by GPT-3.5 align with our human-generated labels—which serve as the ground truth—we subsequently incorporated both the example and GPT-3.5's explanatory rationale into the prompt architecture for the few-shot learning experiment. Finally, we redeploy the zero-shot prompts in the fine-tuning phase to facilitate a comprehensive comparison between zero-shot and fine-tuned model performances. 

\subsection{LLM Fine-Tuning}

Pretraining Large Language Models (LLMs) on extensive corpora enables them to perform competently across a wide range of tasks with minimal examples, often achieving results that rival those of fine-tuned transformer models \citep{brown2020language}.
Specifically, in contexts where the LLM has not been sufficiently trained on task-relevant data, supervised fine-tuning can offer advantages. This involves supplementing the model with an additional dataset of labeled task-specific examples and selectively updating a subset of its weight parameters \citep{wei2022flan, NEURIPS2022_instructGPT}.
However, while effective, fine-tuning such extensive models, particularly those with tens to hundreds of billions of parameters, can be computationally intensive, often requiring large-scale GPU clusters. However, recent advancements have made it feasible to fine-tune these models on single-GPU systems by employing techniques such as 4-bit or 8-bit quantization and adding lower-rank adapter layers to the original architecture \citep{dettmers2023qlora}.

In GPT-3.5, the fine-tuning process bridges the generalized learning acquired from pre-training and the specialized learning required for domain-specific tasks. OpenAI's GPT-3.5 architecture permits fine-tuning via its specialized Application Programming Interface (API), encompassing a multi-step workflow. Initially, the procedure necessitates the preparation of domain-specific datasets, generally constituting labeled instances. The transformation of this data involves segregating the input into three distinct components as mandated by the API. The first segment comprises the system prompt, articulating the overarching task instruction (e.g., Definition, Steps, Examples, etc.). Subsequently, the second segment encapsulates the user prompt, laden with domain-specific data and the instruction that requires labeling. The final segment incorporates the assistant prompt, directly indicating the target labels. To make sure the uploaded data adheres to the requested format, OpenAI provides a function to check for compatibility\footnote{Additional information for data preparation and analysis for chat model fine-tuning can be found here: \href{https://cookbook.openai.com/examples/chat_finetuning_data_prep}{OpenAI Cookbook}}.
.

Acting as a facilitative mechanism, the API enables users to delineate the training configuration, offering the liberty to customize hyperparameters, such as the number of epochs. Once the configuration is established and data uploaded, the API initiates the fine-tuning process. The model parameters are then updated iteratively to minimize the loss on the fine-tuning dataset. However, the inner workings of the fine-tuning process remain somewhat opaque, limiting interpretability and potential improvements. Post-fine-tuning, the model is evaluated on a held-out dataset to ascertain its performance on the target task. The fine-tuned model can then be deployed for the desired application. 

We employed a combination of techniques to achieve efficient adaptation for fine-tuning the Open Source Models. Low-Rank Adaptation (LoRA) significantly reduces the number of trainable parameters by introducing low-rank matrices into each layer that capture task-specific adjustments \citep{hu2021lora}. Additionally, 4-bit quantization compresses the pre-trained model weights from 32-bit floating-point numbers to a more memory-efficient 4-bit representation, as described in \citet{dettmers2024qlora}. This combination allows us to perform supervised fine-tuning on large models like FLAN-T5-xl \citep{wei2022flan},  ‘oasst-sft-6-LLaMA-30b’ and  ‘LLaMA-2 13b/70b’ \citep{köpf2023openassistant} with better efficiency and potentially faster training times. We selected the xl-version of FLAN due to its enhanced capabilities relative to its smaller counterparts while keeping the computational demands reasonable. We used adapter layers on the Query and Value attention blocks in all three cases. As the training sets for each task are small and all are text classification tasks, we chose $r=16$ and $\alpha=32$ as hyperparameters for the adapter layers added. A lower rank was chosen to avoid overfitting to the training set, while the $\alpha$ was selected to produce a scaling of 2 and give more weight to the output of the adaptive layers and force the LLMs to follow the format used in the training set examples. As for the hyperparameters during training, we chose the default parameters of the Seq2SeqTrainer and SFTTrainer from huggingface \citep{vonwerra2022trl}. 

To fine-tune ‘flan-t5-xl’, we used a single 80GB A100 GPU. The training examples had as input the zero-shot prompt with the coding or labeling guidelines followed by the text to label, and as output the letter that identified the label that should be assigned to the text (i.e: in the Relevance task 'A' for relevant texts or 'B' for irrelevant texts).
For oasst-LLaMA, two 80GB A100 GPUs were needed to fine-tune the model. For the trainset examples, we followed the original prompt format used to fine-tune LLaMA-1 and produce the ‘oasst-sft-6-LLaMA-30b’ model\footnote{\href{https://github.com/LAION-AI/Open-Assistant/blob/main/model/MESSAGE\_AND\_TOKEN\_FORMAT.md}{https://github.com/LAION-AI/Open-Assistant/blob/main/model/MESSAGE\_AND\_TOKEN\_FORMAT.md}}. In the segment designated for the prompter section, we reincorporated the zero-shot prompt and the corresponding text requiring labeling. Conversely, in the assistant section of the prompt, we included the target label to which the text should be mapped to. For LLaMA-2, we needed three 80GB A100 GPUs for the 70b model while only using one for the smaller 13b model. Furthermore, we maintained consistency with previous fine-tuning efforts by adhering to the original prompt format established for the "oasst-sft-6-LLaMA-30b" model. To optimize training efficiency, the input text for both models (LLaMA-1 \& LLaMA-2) was left-truncated at a maximum of 4096 tokens. This combination of prompt design, targeted allocation of computational resources, and adherence to established prompting strategies facilitated effective fine-tuning of the LLaMA models. We trained all three LLMs for three epochs with a batch size of four. To further improve efficiency, we implemented a technique called gradient accumulation, where the model weights were only updated after accumulating gradients from every second batch.

\subsection{Evaluation Metrics}\label{section:eval}
We computed average accuracy (i.e. percentage of correct predictions), that is, the number of correctly classified instances over the total number of cases to be classified, using trained human annotations as our gold standard and considering only texts that both annotators agreed upon. Second, in applicable cases, we computed intercoder agreement, measured as the percentage of instances for which both annotators in a given group report the same class.

For the internal evaluation of the fine-tuning process, we employed additional metrics to garner more profound insights into the models' learning and generalization across classes. Specifically, we calculated each class's precision, recall, and F1 score to ensure that the fine-tuning process was comprehensive and aimed at enhancing performance across all classes. Precision is the ratio of correctly predicted positive observations to the total predicted positives, providing insight into the models' ability to identify positive instances accurately. On the other hand, recall is the ratio of correctly predicted positive observations to all observations in actual class, shedding light on the models' capability to identify all possible positive instances. The F1-score is the weighted average of precision and recall, thereby balancing the two metrics, especially in cases where one may have more significance than the other. Employing these metrics facilitated a robust evaluation, ensuring that the models were not biased towards the majority class and that the fine-tuning process effectively enhanced the models’ performance across all classes.

\section{Results}
All results in this paper extends a previous study which compared GPT-3.5's zero-shot annotation performance with that of MTurk \citep{gilardi2023chatgpt}. We rely on extended datasets (n = 9,777), which include tweets and news articles that were collected and annotated manually on the discourse around content moderation \citep{alizadeh2022content}, as well as a new sample of tweets posted in 2023 to address the concern that LLMs might be merely reproducing texts that could have been part of their training data. While the previous study used only GPT-3.5 for text classification, our analysis conducts the same classifications using GPT-4 as well as two open-source LLMs (LLaMA and FLAN), using the same codebook that was originally constructed for the research assistants and MTurkers (see \ref{appendix:codebook_zeroshot}).

\subsection{Choosing the training approach: Zero-Shot versus Few-Shot}
Probably the first decision with respect to using LLMs for text annotation is whether to first manually annotate a subset of data and use it for few-shot learning or just directly proceed with a LLM in a zero-shot setting (see \citep{brown2020language} for more background). Moreover, even if a researcher decides to have some manually annotated data, then the question is whether to use crowd-workers, or to recruit expert research assistants. The later question has already been answered in a previous study \citep{gilardi2023chatgpt}, in which the authors showed that GPT-3.5 outperforms crowd workers for several annotation tasks, including relevance, stance detection, topic modeling, and frame detection. Across the four datasets and a total of 12 annotation tasks, the zero-shot accuracy of GPT-3.5 exceeds that of crowd workers by about 25 percentage points on average. 

For the purpose of answering the question of whether to go zero-shot or few-shot, before comparing these approaches, we would like to highlight the necessity of measuring the accuracy of LLMs with a priori annotated data. In fact, no matter how well GPT-3.5 or other LLMs have been reported to perform across various datasets and text annotation tasks, whenever a researcher is using a new dataset or need to implement a new annotation task, it is recommended to measure the accuracy of LLMs on a small subset of manually annotated data. The size of the test set varies in different papers (e.g. \citep{gilardi2023chatgpt} and \citep{ziems2023can}) and its somehow arbitrary. However, we recommend that a human expert manually annotate at least 100 data points.

\begin{figure*}[t]
  \centering
  \includegraphics[width=0.80\linewidth]{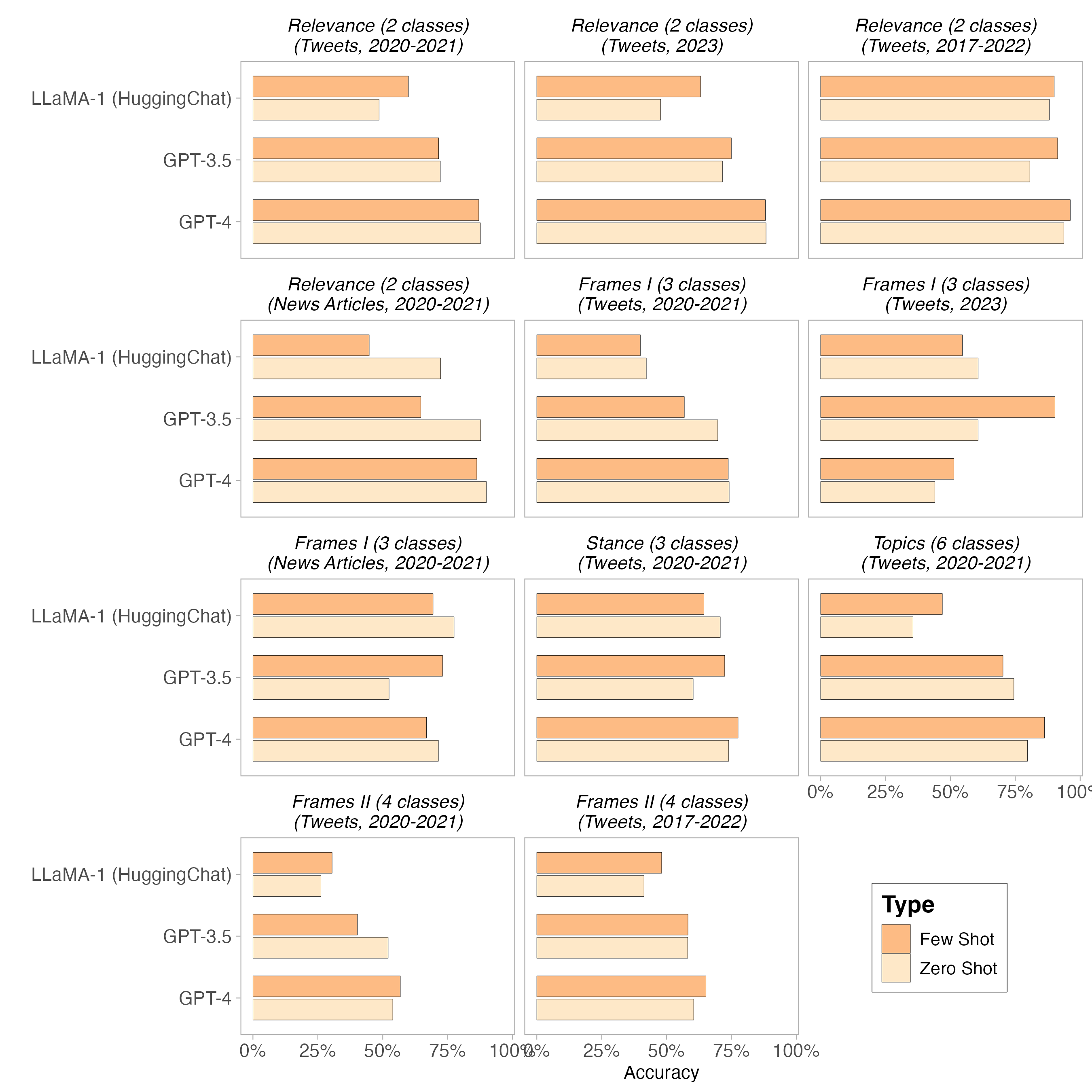}
 \caption{Comparing zero- and few-shot text annotation of GPT-3.5, GPT-4, and LLaMA-1 (HuggingChat). The x-axis shows the accuracy. The y-axis displays the two models grouped by the model configuration, including Zero-Shot and Few-Shot. Facets represent distinct tasks and/or datasets for evaluating model configurations.}
\label{figure:results_zero_few}
\end{figure*}

To understand whether it is safe for researchers to seamlessly use a LLM for text annotation in a zero-shot setting or not, here we extend the previous analysis of \citep{gilardi2023chatgpt} to include few-shot learning. We conduct chain of thought (CoT) prompting \citep{wei2022chain}. This few-shot approach involves providing LLMs with question and step-by-step reasoning answer examples. In addition, we include results obtained from GPT-4 and HuggingChat (which uses LLaMA-1) as well. We chose HuggingChat due to its popularity and ease of use. The corresponding prompts are reported in \ref{appendix:codebook_fewshot_cot}. The results are illustrated in Figure \ref{figure:results_zero_few}. Overall, in Figure \ref{figure:results_zero_few}, we can see that the few-shot results are mixed, with some tasks slightly benefiting from few-shot learning, some performing lower, and some with no difference. For GPT-4 and GPT-3.5, we see performance gain for few-shot learning in 6 tasks and performance reduction in 5 tasks, though it varies across tasks. As for LLaMA-1 (HuggingChat), we see performance gain for few-shot learning in 4 tasks and performance reduction in 7 tasks.

With respect to what explains the mixed performance of few-shot learning across various text annotation tasks, we could not find any conclusive pattern. For example, let's consider the number of classes as the measure for the complexity of the classification tasks. We see that the LLaMA-1 (HuggingChat) and GPT-4 benefited from few-shot learning in majority of more complex tasks, including topic modeling (6 classes), and framing detection (4 classes). However, GPT-3.5 saw a performance reduction in these three tasks. On the other hand, for the less complex tasks of relevance (first two plots from the top left in Figure \ref{figure:results_zero_few}), we see that LLaMA-1 (HuggingChat) experienced significant performance gain from few-shot learning, but GPT-4 and GPT-3.5 results are mixed. As another example, exploring different types of datasets (tweets vs. news articles), we see that all GPT-4, GPT-3.5, and HuggingChat are showing less accuracy for few-shot learning in the relevance tasks, but in the framing detection task, the results are mixed, with GPT-3.5 benefiting from few-shot learning and GPT-4 and HuggingChat losing performance from it. 

\begin{tcolorbox}[colback=white,sharp corners,boxrule=0.5pt]
\begin{center}
\textbf{Choosing a Training Approach: Zero- vs. Few-Shot}                   
\end{center}

\textbf{Advantages:}

\textit{Zero-Shot:} Off-the-shelf usage; no annotation cost; academic benchmarks on performance gain with few-shot learning are inconclusive.\newline
\textit{Few-Shot:} Possibility of in-context learning with minimal examples; might lead to performance improvement.
\vspace{3mm}

\textbf{Findings:}
In our tests, few-shot results are mixed, some tasks benefited from it and some lost performance. No explicit pattern with respect to task complexity, model selection, and data type. 
\vspace{3mm}

\textbf{Advice:}
Always manually annotate 100-250 data points to measure the accuracy of LLMs, specailly If working on a new dataset or task. We do not recommend spending much money and time on few-shot learning. 
\end{tcolorbox}

\subsection{Temperature Setting: Higher versus Lower}

Another important decision that a researcher should make about using LLMs for text annotation is about the value of the temperature parameter. Both GPT-3.5 and LLaMA-1 (HuggingChat) have a temperature parameter which controls the degree of randomness, and thus the creativity, of the output. A higher temperature will result in more diverse and unexpected responses, while a lower temperature will result in more conservative and predictable responses. The default temperature value is 1.0 for GPT-3.5 and 0.9 for HuggingChat. Previous research showed that a lower temperature value may be preferable for text annotation tasks, as it seems to increase consistency without decreasing accuracy \citep{gilardi2023chatgpt}. Here, we extend the previous research results by assessing the effect of a lower temperature in LLaMA-1 (HuggingChat). Similar to \citep{gilardi2023chatgpt}, we set the temperature at its default value and 0.2 and compare the outputs with respect to accuracy and intercoder agreement. We conducted two sets of annotations for each temperature value to compute LLM's intercoder agreement. 

Our results demonstrate that lower temperature settings significantly enhance intercoder agreement, underscoring the deterministic and repeatable nature of the annotations. For instance, GPT-3.5's average intercoder agreement increased from 91.7\% to 97.6\% when the temperature was reduced from 1 to 0.2 in the zero-shot setting and from 92.3\% to 95.4\% in the few-shot setting. Similarly, for LLaMA-1 (HuggingChat), the agreement surged from 46.7\% to 84.8\% in the zero-shot and from 47.1\% to 83.1\% in the few-shot settings when the temperature was lowered from 0.9 to 0.2. These substantial improvements in intercoder agreement with lower temperatures provide a compelling case for their use in ensuring more deterministic and reliable LLM annotations. Moreover, our examination of accuracy values further supports the preference for lower temperature settings. Notably, the accuracy in the task of 'Stance' classification on Tweets from 2020-2021 increased remarkably from 53.7\% to 70.7\% as the temperature lowered from 0.9 to 0.2. Additionally, in the 'Relevance' classification for News Articles from 2020-2021, we observed a significant accuracy boost from 56.6\% to 72.3\% when the temperature setting was reduced. These examples underscore that a more deterministic approach, achieved by lowering the temperature, improves consistency and enhances the overall quality of results in various classification tasks, affirming the efficacy of employing lower temperature settings for improved LLM performance.

\begin{figure*}[h]
  \centering
  \includegraphics[width = 0.8\linewidth]{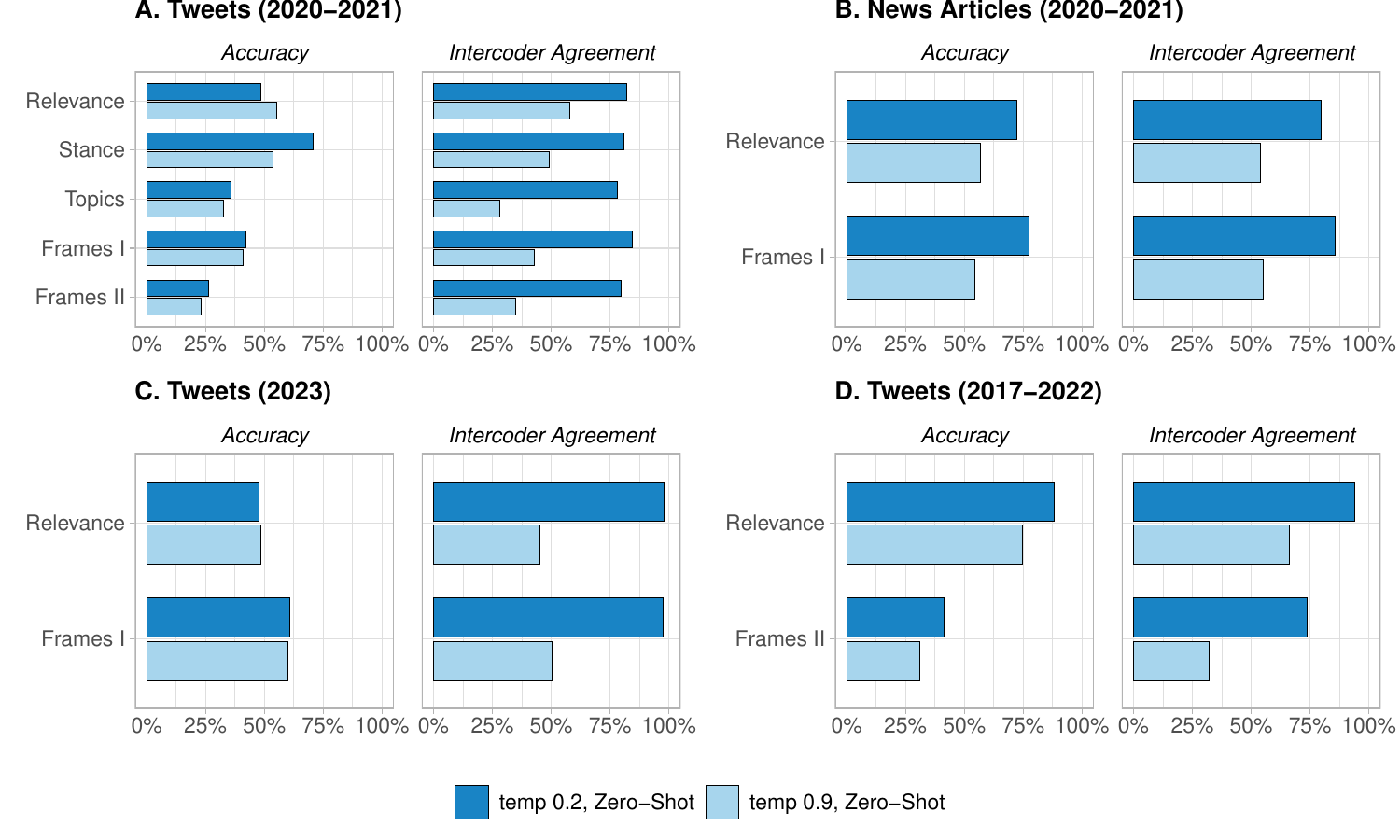}
  \caption{Analyzing the effect of LLaMA-1 (HuggingChat)'s temperature parameter on accuracy and intercoder agreement in text annotation tasks.}
  \label{temperature_hugg}
\end{figure*}	

\begin{figure*}[h]
  \centering
  \includegraphics[width = 0.8\linewidth]{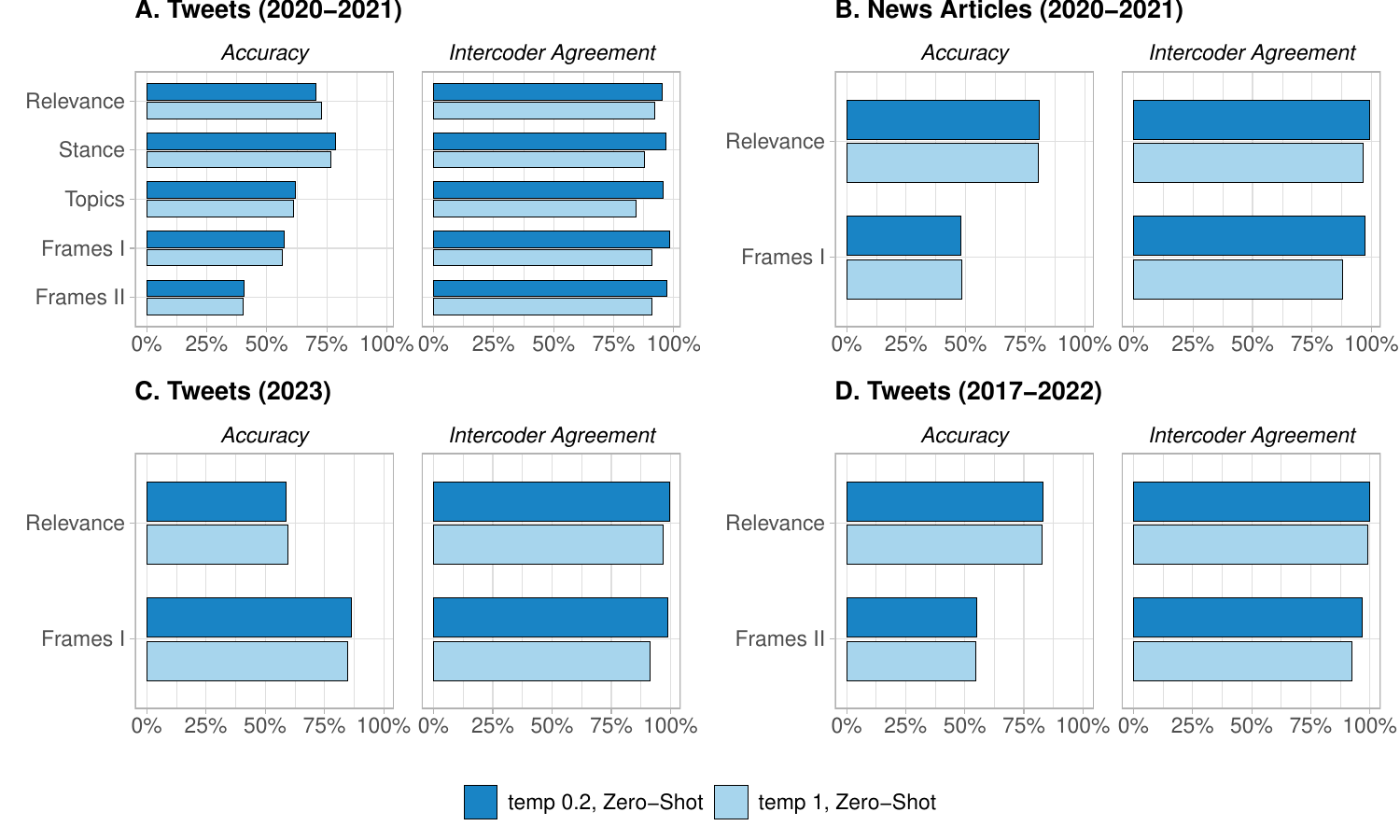}
  \caption{Analyzing the effect of GPT-3.5's temperature parameter on accuracy and intercoder agreement in text annotation tasks.}
  \label{temperature_gpt}
\end{figure*}

Across the four datasets, we report HuggingChat’s zero-shot performance for two different metrics: accuracy and intercoder agreement (Figure \ref{temperature_hugg}). Accuracy is measured as the percentage of correct annotations (using our trained annotators as a benchmark), while the intercoder agreement is computed as the percentage of tweets that were assigned the same label by two different annotators (research assistant, crowd-workers, or GPT-3.5's runs). Figure \ref{temperature_hugg} shows that while the accuracy and intercoder agreement are, on average, lower than those reported for GPT-3.5 in \citep{gilardi2023chatgpt}, the pattern for the effect of temperature is the same. Across all four datasets and eleven annotation tasks, decreasing the temperature significantly improved the intercoder agreement scores without decreasing the accuracy. The only exception is for the relevance task in Tweet (2020-2021) dataset (first row in the top left plot in Figure \ref{temperature_hugg}) which decreasing the temperature to 0.2 increased the intercoder agreement but led to reduction in accuracy. Interestingly, the GPT-3.5 results in \ref{temperature_gpt} shows the same pattern for this particular task and dataset.

\begin{tcolorbox}[colback=white,sharp corners,boxrule=0.5pt]
\begin{center}
\textbf{Temperature Setting: High vs. Low}                   
\end{center}

\textbf{Advantages:}

\textit{Low Temperature:} Less randomness; less creative answers; more deterministic output.\newline
\textit{High Temperature:} More randomness, more creative answers, more variations in outputs.
\vspace{3mm}

\textbf{Findings:}
Our analyses show that in almost all annotation tasks, lower temperature setting for GPT-3.5 and HuggingChat increases the intercoder aggreemnt rate without decreasing the accuracy. 
\vspace{3mm}

\textbf{Advice:}
We recommend setting the temperature parameter at zero (no randomness). 
\end{tcolorbox}

\subsection{Model Selection: Proprietary versus Open-Source LLMs}

Proprietary closed-source LLMs such as ChatGPT and Bard (Gemini) are more convenient and safe to use for general audiences due to their heavy fine-tuning to align with human preferences \citep{Spirling:2023aa}. Although the training methodology is straightforward and simple, the extensive computational demands have restricted the creation of LLMs to a select few. That is why none of the open-source LLMs such as BLOOM, LLaMA-1 and Falcon could made a suitable substitutes for closed-source LLMs \citep{touvron2023llama}. More recently, responding to this demand, LLaMA-2 was introduced, which is a family of pretrained and fine-tuned LLMs at scales up to 70B parameters \citep{touvron2023llama}. LLaMA-2 evaluations showed it outperforms LLaMA-1, Falcon, and MPT in standard academic benchmarks including commonsense reasoning, world knowledge, reading comprehension, and math, and performs on par with GPT-3.5 in math and popular aggregated benchmarks but fall short in Python code writing benchmarks \citep{touvron2023llama}.

\begin{figure}[hb!t]
  \centering
  \includegraphics[width=0.90\linewidth]{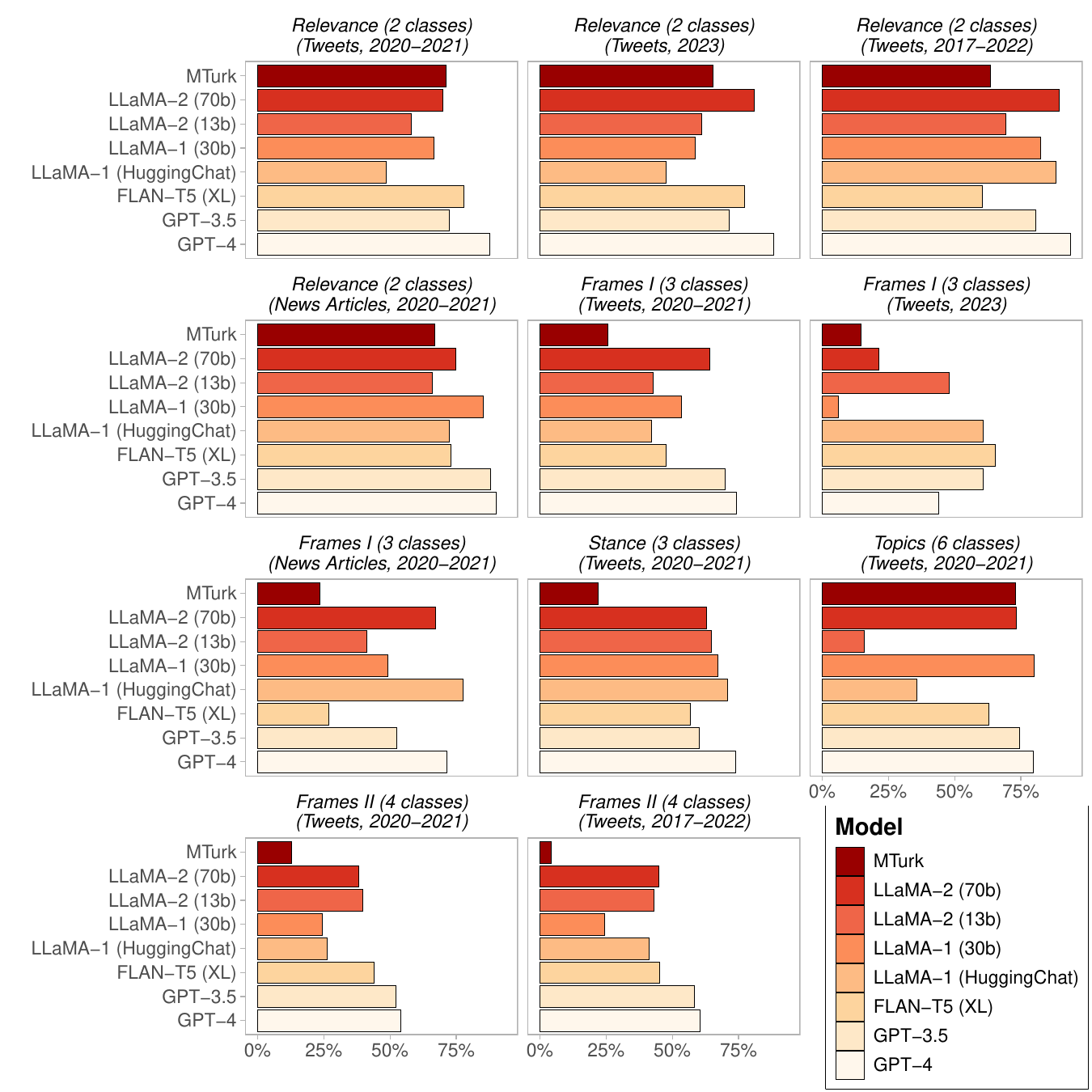}
 \caption{Accuracy of GPT-3.5, GPT-4, open-source LLMs, and MTurk. Accuracy means agreement with trained annotators. Bars indicate average accuracy, while whiskers range from minimum to maximum accuracy across models with different parameters and/or prompts (zero vs few shot).}
\label{figure:results_open_gpt_mturk}
\end{figure}

Text annotation has always been costly. Although previous findings showed that GPT-3.5 outperforms Amazon Mechanical Turk crowd-workers (MTurker) and costs almost thirty times cheaper \citep{gilardi2023chatgpt}, it is not free of charge. Hence, it is tempting for researchers to explore the extent to which open-source LLMs are capable for text annotation tasks. In addition to cost-effectiveness, open-source LLMs are increasingly recognized for their transparency, reproducibility, and enhanced data protection features \citep{Spirling:2023aa,Ollion:2024aa}. However, the academic benchmarks reported above lack the text annotation tasks, especially that of political text. To assess how well open-source LLMs perform in text annotation tasks, we compare GPT (3.5 \& 4) results with those of LLaMA (1 \& 2) and FLAN (T5 XL). Considering the reported training data size, scaled-up parameters, and reading comprehension benchmarks, we expect the Llama-2 (70b) model to perform well on our text annotation tasks.

Figure \ref{figure:results_open_gpt_mturk} compares the text annotation accuracy of GPT-4, GPT-3.5, LLaMA-1 (HuggingChat), LLaMA-1 (30b), LLaMA-2 (13b), LLaMA-2 (70b), FLAN-T5 (XL), and MTurkers. Six observations stand out in this analysis: (1) no LLM outperforms others across all 11 annotation tasks; (2) only GPT-4 and GPT-3.5 outperform MTurkers in all 11 annotation tasks; (3) among the open-source LLMs, the best performing one in terms of the number of tasks that it outperforms MTurkers is LLaMA-2 (70b), with 9 out of 11 annotation tasks that it outperforms the crowd-workers; (4) among the open-source LLMs, the worst performing one in terms of the number of tasks that it outperforms MTurkers is LLaMA-2 (13b), with only 6 out of 11 annotation tasks that it outperforms the crowd-workers; (5) among the open-source LLMs, the best performing one in terms of outperforming GPT-3.5 is LLaMA-2 (70b), with 5 out of 11 annotation tasks it outperforms GPT-3.5; (6) across the 9 annotation tasks that are related to datasets published before 2023, all open-source LLMs outperform crowd-workers where the number of classes is three and greater. In other words, when the data falls within an open-source LLM's cutoff date and the annotation task is not a binary classification, all open-source LLMs perform better than MTurkers. 

Overall, these findings underscore that while open-source LLMs are not consistently the superior choice, they generally outperform crowd-sourced annotations and are approaching the performance levels of GPT 3.5. Even an out-of-the-box tool such as HuggingChat, which looks and works very much like ChatGPT, outperforms Amazon Mechanical Turk crowd-workers in 8 out of all annotation tasks and performs close enough to GPT 3.5 in 6 out of 11 annotation tasks. While there is no universal answer to the question of what is the best open-source LLM for political text annotation, and the best performing LLM varies across dataset, task, and model size, our results show that a collection of open-source LLMs could almost perform on par with GPT-3.5. Therefore, we recommend to (1) use open-source LLMs for text annotation in social sciences research; and (2) compare the performance of 2-3 open-source LLMs, such as LLaMA-2 (70b), FLAN-T5 (XL), and LLaMA-1 (HuggingChat), and pick the best performing one. 

\begin{tcolorbox}[colback=white,sharp corners,boxrule=0.5pt]
\begin{center}
\textbf{Model Selection: GPT-3.5/4 vs. Open-Source LLMs}                   
\end{center}

\textbf{Advantages:}

\textit{GPT-3.5/4:} Off-the-shelf usage; more convenient and safe to use; heavy fine-tuning to align with human preferences.\newline
\textit{Open-Source LLMs:} No cost (GPT-3.5, and especially GPT-4 can become expensive for researchers without large research budgets); more transparency; ethical way to do research due to data privacy concerns; more reproducibility.
\vspace{3mm}

\textbf{Findings:}
Our results show that while open-source LLMs are not consistently the superior choice, they generally outperform crowd-sourced annotations and are approaching the performance levels of GPT 3.5 (ChatGPT). 
\vspace{3mm}

\textbf{Advice:}
Use both LLaMA-2 (70b) and HuggingChat if resources allow for running heavy-size models, and pick the best-performing model. Use HuggingChat (LLaMA-1) if high-performance computing is not available. Do not use LLaMA-2 (13b). 
\end{tcolorbox}


\subsection{Annotation Size: How much annotation is enough for fine-tuning?}

In this section, we are interested in testing the effect of fine-tuning on the performance of both closed- and open-source LLMs. Advanced large language models (LLMs), such as GPT-4 and LLaMA-2, often demonstrate new capabilities and can learn from context with minimal examples, enabling them to perform complex tasks \citep{wei2022chain, shen2024hugginggpt}. However, fine-tuning these models is still necessary to unlock their full potential for creative and specialized tasks, aligning their performance with human preferences \citep{yang2023bigtrans, schick2024toolformer}. Here, we would like to answer three important questions: (1) Would LLM exhibit performance gain in text annotation tasks when they get fine-tuned with human expert annotated data?; (2) If fine-tuning improves the performance of LLMs in text annotation accuracy, how big the training data should be?; and (3) Does the effect of fine-tuning on LLMs' text annotation accuracy varies between close- and open-source LLMs? 

Several factors may influence the efficacy of fine-tuning LLMs, including but not limited to 1) pretraining conditions; and 2) fine-tuning conditions. Pretraining factors include the size of the LLM and the volume of pretraining data, which are critical in determining the quality of the representation and knowledge encoded in the pretrained LLMs. On the other hand, fine-tuning conditions such as the nature of the downstream task, the size of the fine-tuning dataset, and the specific fine-tuning methodologies employed can significantly impact the extent of knowledge transfer to the targeted task \citep{zhang2024scaling}. Prior research has extensively investigated the scaling of LLM pretraining or training from scratch \citep{hoffmann2022training} as well as the development of advanced methods for fine-tuning \citep{he2021towards}. However, the issue of whether and how the fine-tuning of LLMs scales with the fine-tuning conditions has been largely overlooked.

In this section, we are interested to explore the effect of LLM selection, LLM size, and the size of the fine-tuning data on the accuracy of LLMs on our running text annotation text. A recent study, based on three downstream tasks on translation and summarization, finds that the augmentation of the LLM model's size exerts a more substantial influence on fine-tuning compared to increasing the size of fine-tuning data \citep{zhang2024scaling}. Furthermore, their results show that the effectiveness of fine-tuning varies across tasks and datasets, making the selection of the best fine-tuning approach for a particular downstream task less definitive. Considering these recent findings, we should expect to see a similar pattern, in which the performance gain from fine-tuning is being dependent on the text annotation task, model, model size, and the size of fine-tuning data.  

\begin{figure}[t]
  \centering
  \includegraphics[width=0.95\linewidth]{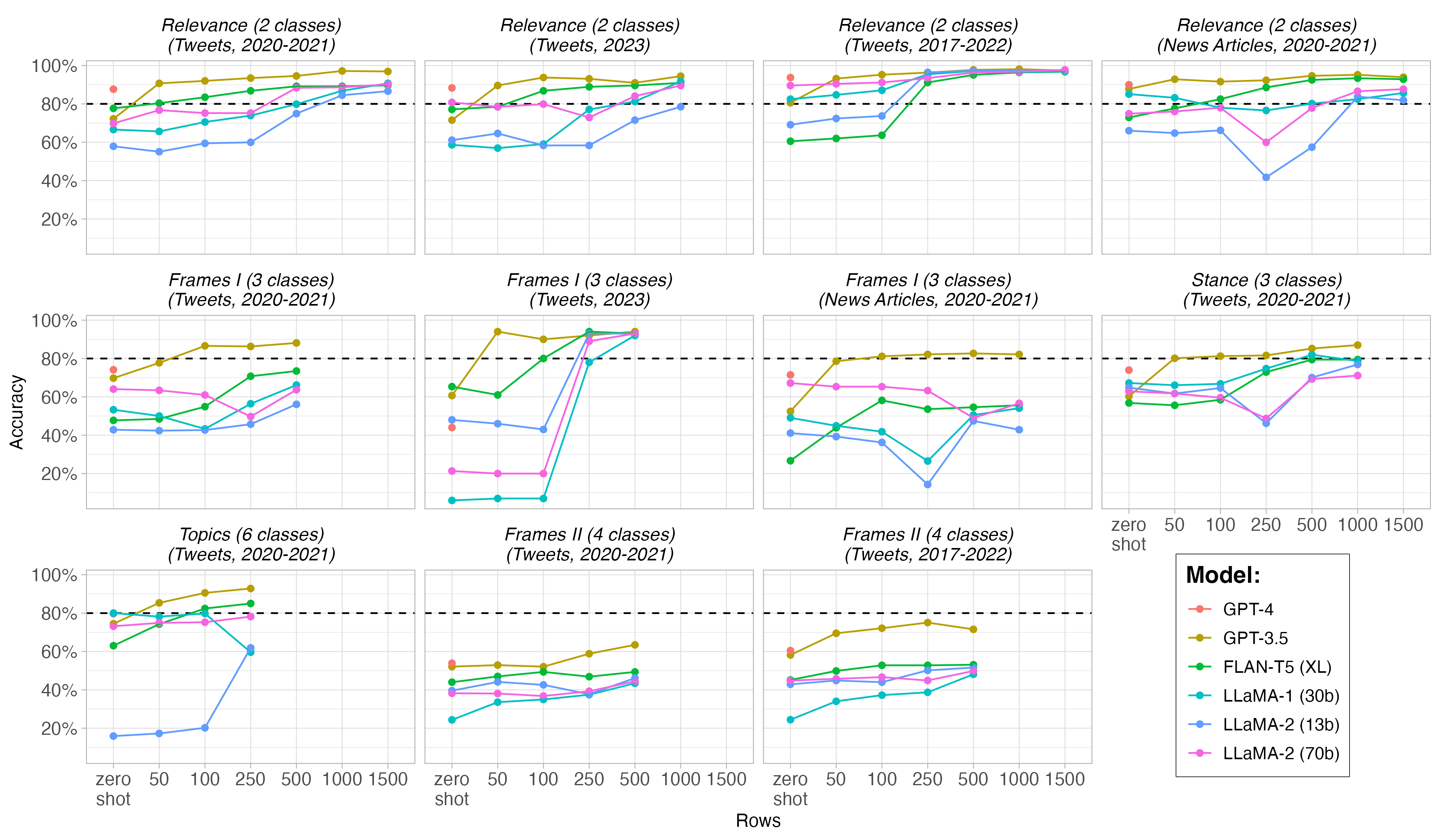}
 \caption{Performance (accuracy) of GPT-3.5, LLaMA-1, LLaMA-2, and FLAN-T5 (XL), as a function of the training data size for fine-tuning. The x-axis shows different sizes of training datasets, ranging from zero-shot (no fine-tuning) to 50, 100, 250, 500, and 1,000 rows used for fine-tuning the models. The y-axis displays the accuracy of the models in percentages. Facets represent distinct tasks and/or datasets for evaluating the models. Pink dots represent zero-shot GPT-4 accuracy for the sake of comparison.}
\label{figure:results_ft}
\end{figure}

Figure \ref{figure:results_ft} compares fine-tuning accuracy across five LLMs (LLaMA-1 (30b), LLaMA-2 (13b), LLaMA-2 (70b), FLAN-T5 (XL), and GPT 3.5), 11 text annotation tasks, between 4 to 7 different sizes of fine-tuning data (corresponding F1-scores are reported in Fig \ref{results_f1} in the Appendix). Several observations stand out in Figure \ref{figure:results_ft}. First, with the exception of LLaMA-1 (30b), we see a general trend in other LLMs, in which the accuracy and F1-score of models improves with the size of fine-tuning data. Second, one of our salient findings in Figure \ref{figure:results_ft} and Figure \ref{results_f1} centers on the capabilities of FLAN-T5 XL. Our analysis reveals that, when fine-tuned, FLAN-T5 (XL) matches or even surpasses the zero-shot accuracy and F1-score of ChatGPT across all tasks and datasets, with a singular exception. It falls short in classifying the Problem-Solution frames in the 2023 tweets.

The third, and arguably the most compelling, finding in Figure \ref{figure:results_ft} and Fig \ref{results_f1} centers on the differential rates and magnitudes of improvement observed across models during the fine-tuning process. ChatGPT demonstrates remarkable performance gains, even when fine-tuned with a minimal dataset of just 50 instances. Specifically, it registers an average accuracy increase of 15.7 \%, which further escalates to 19.1 \% when the training set comprises 100 cases. In a stark contrast, open-source models such as FLAN-T5 (XL) and LLaMA-1 exhibit a more incremental progression in performance. Intriguingly, the LLaMA-1 model initially sees a dip in accuracy on certain tasks but tends to recover and improve as the dataset expands to around 250 instances. By this point, both FLAN-T5 (XL) and LLaMA-1 come close to matching the zero-shot accuracy levels achieved by ChatGPT. It is worth noting that the average accuracy gains for FLAN-T5 (XL) stand at 7.7 \% and 12.4 \% when fine-tuned with 50 and 100 instances, respectively. This underscores the point that open-source models, too, stand to benefit significantly from fine-tuning. 

These findings illuminate the complex dynamics at play in the fine-tuning of LLMs for text annotation tasks. They highlight the variable performance across different models and tasks, the rapid yet plateauing gains for commercial models like ChatGPT, and the more gradual but sustained improvements for open-source models. Next, fine-tuning is particularly helpful when using GPT-3.5 Turbo, even with as little as 50 rows of training data. The results are mixed for open-source models like FLAN-T5 (XL) and LLaMA-1. FLAN-T5 (XL) generally benefits from fine-tuning, while LLaMA-1's performance is inconsistent. However, as shown in Figure \ref{figure:results_ft}, increasing the amount of training data does improve LLaMA-1's performance over its zero-shot capabilities, just like for all other models.

In our study, we observed that the cost of fine-tuning commercial models like ChatGPT is quite reasonable. The total expenditure for fine-tuning all our ChatGPT models was only 311 \$, with an additional 34 \$ for the evaluation of these models. Specifically, fine-tuning GPT-3.5 costs merely 1.2 \$ for 100 rows across three epochs, with subsequent usage costs being only 16 Cents per 100 rows. These figures illustrate that even when relying on commercial models, fine-tuning is an economically viable option, especially when compared to the costs of zero-shot usage. While open-source LLMs can offer significant cost savings when suitable infrastructure is available, fine-tuning commercial models like ChatGPT remains an affordable and efficient alternative for those without access to such resources.

In summary, our empirical findings robustly advocate for fine-tuning as the primary strategy for enhancing classification accuracy across varying tasks and models. Although zero-shot and few-shot learning paradigms may offer utility under specific circumstances, fine-tuning emerges as the most consistently effective approach. However, it is important to note that the efficacy of fine-tuning is not universally high across all task complexities. Specifically, the availability of substantial, high-quality training datasets becomes imperative for achieving optimal performance levels for intricate classification tasks. Despite fine-tuning, it remains possible that the model's performance may not meet the thresholds required for specific specialized applications. This is particularly relevant as more labeled datasets become available, making fine-tuning a practical choice for those aiming to optimize classification tasks.

\begin{tcolorbox}[colback=white,sharp corners,boxrule=0.5pt]
\begin{center}
\textbf{Annotation Size: How much annotation is enough for fine-tuning?}                   
\end{center}

\textbf{Advantages:}

\textit{50-100 Manual Annotation:} Less cost; less time consuming.\newline
\textit{250-500 Manual Annotation:} Potential for higher performance gain.
\vspace{3mm}

\textbf{Findings:}
In general, LLMs' accuracy increases with the size of fine-tuning data. FLAN-T5 (XL) matches or even surpasses GPT-3.5's zero-shot performance across all tasks and datasets (except one task). Open-source LLMs and GPT-3.5 differ in the optimal number of required fine-tuning data. 
\vspace{3mm}

\textbf{Advice:}
Our empirical findings robustly advocate for fine-tuning as the primary strategy for enhancing classification accuracy. If using GPT-3.5, 50 annotated data points are enough. If using an open-source LLM, go with 250.
\end{tcolorbox}

\section{Conclusion}

Automated classification of short and long text is central to a growing number of research questions in the social sciences. Previous research advocated for using supervised machine learning methods over dictionary-based approaches and provided best practices for human annotation of text \citep{barbera2021automated}. However, the emergence of large language models (LLM) and their ability to outperform crowd-workers in text annotation and yielding acceptable accuracy compared to human expert evaluation \citep{gilardi2023chatgpt} provide researchers with new opportunities to skip the crowd-sourcing or even training their own supervised machine learning models for text classification. Nevertheless, in the rush to take advantage of these opportunities, one can easily neglect to consider crucial questions and underestimate the implications of certain choices. 

In this paper, we have tried to walk the researchers through critical decisions they need to make for using LLMs in text annotation tasks (e.g. relevance, topic detection, and framing detection) and provided them with some practical advice backed by our empirical results. Our most surprising finding is the substantial effect of fine-tuning on increasing LLMs text annotation performance. We demonstrate that open-source LLMs such as LLaMA-1, LLaMA-2, and FLAN represent a competitive alternative for text annotation tasks, exhibiting performance metrics that generally exceed those of Amazon Mechanical Turk crowd-workers and rival those of GPT 3.5 (ChatGPT). An important appeal of open-source LLMs is that they offer considerable cost advantages. While ChatGPT provides substantial cost-efficiency, being about thirty times more affordable per annotation compared to MTurk \citep{gilardi2023chatgpt}, open-source LLMs surpass this by being freely available. This constitutes a significant improvement in the accessibility of such models, extending their reach to a broader range of researchers irrespective of financial constraints.

Open-source LLMs present benefits that go beyond cost-efficiency. One key advantage is that they help reduce reliance on proprietary models operated by for-profit companies, which may conflict with research ethics and the reproducibility standards \citep{Spirling:2023aa,Ollion:2024aa}. Furthermore, open-source LLMs provide distinct benefits for data protection, as they are designed in such a way that data do not need to be shared with any third-party entities \citep{van2023chatgpt}. This feature ensures that sensitive information remains secure and confidential because it is not sent to or stored by an external party. The elimination of data sharing in open-source LLMs provides an extra layer of protection against potential data breaches or unauthorized access. This feature becomes especially beneficial in scenarios where sensitive data is involved, such as in the legal or medical fields, where confidentiality is of utmost importance \citep{ray2023chatgpt, paul2023digitization, murdoch2021privacy}, but also in social science research involving data protected under the European Union's General Data Protection Regulation (GDPR), or covered by non-disclosure agreements (NDAs).

We conclude with four general pieces of advice for text analysts: (1) manually annotate 250-500 data points and use half for fine-tuning and half for accuracy testing; (2) use fine-tuned open-source LLMs for text annotation due to their cost-effectiveness, transparency, and reproducibility; (3) always validate the output of LLMs by human expert evaluation; (4) run LLMs at least twice per task and report average accuracy and intercoder agreement; and (5) set the temperature of GPT or LLaMA models at zero or a low value to get higher intercoder agreement without loss in accuracy. Nonetheless, we emphasize that our results are not generalized to all text annotation tasks and datasets. Taking vital decisions discussed in this paper, researchers should perform their own testing, especially when working with different datasets or dealing with new annotation tasks. 



\paragraph{Acknowledgments}

We thank Darya Zare, Fabio Melliger, Mohammadamin Alizadeh, Paula Moser, Mahdis Abbasi, Sophie van IJzendoorn, and Zahra Baghshahi for excellent research assistance.

\paragraph{Funding Statement}

This project received funding from the European Research Council (ERC) under the European Union's Horizon 2020 research and innovation program (grant agreement nr. 883121).

\paragraph{Data Availability Statement}

As part of our commitment to transparency and reproducibility, we have made all the necessary files available to replicate the analyses presented in this manuscript. The replication package includes datasets, jupyter Python notebook for using and fine-tuning open-source LLMs, and additional supplementary materials used in our study. The replication files can be accessed at the following URL: \url{https://osf.io/ctgqx/}.

\paragraph{Competing Interests}

None.


\bibliographystyle{apsr}
\bibliography{example.bib}

\newpage

\appendix
\renewcommand{\thesection}{S\arabic{section}}
\setcounter{figure}{0} \renewcommand{\thefigure}{S\arabic{figure}}
\setcounter{table}{0} \renewcommand{\thetable}{S\arabic{table}}

\section{Zero-Shot Annotation Codebook} \label{appendix:codebook_zeroshot}

\subsection{Dataset 1: Content Moderation Tweets (2020-2021)}
\subsubsection{Task 1: Relevance \newline}
“Content moderation” refers to the practice of screening and monitoring content posted by users on social media sites to determine if the content should be published or not, based on specific rules and guidelines.

I will ask you to classify a text as relevant or irrelevant to the content moderation:

A: Text is RELEVANT if it includes: social media platforms’ content moderation rules and practices, censorship, governments’ regulation of online content moderation, and/or mild forms of content moderation like flagging, shadowbanning, or account suspension.

B: Text is IRRELEVANT if they do not refer to content moderation, as defined above. This would include, for example, a tweet by Trump that Twitter has labeled his tweet as “disputed”, or a tweet claiming that something is false.

Now, is the following text relevant or irrelevant to content moderation?

[Paste a tweet here and remove the brackets]

\subsubsection{Task 2: Problem/Solution Frames \newline}

“Content moderation” refers to the practice of screening and monitoring content posted by users on social media sites to determine if the content should be published or not, based on specific rules and guidelines.

I will ask you to classify a text as describing content moderation as a problem, as a solution, or neither:

A: Text describes content moderation as a PROBLEM if they emphasize negative effects of it, such as restrictions to free speech, censorship, or the biases that can emerge from decisions regarding what users are allowed to post.

B: Text describes content moderation as a SOLUTION if they emphasize positive effects of it, such as protecting users from harmful content such as hate speech, misinformation, illegal adult content, or spam. 

C: Text describes content moderation as NEUTRAL if they do not emphasize negative or positive effects of content moderation. For example if they simply report on the content moderation activity of social media platforms without linking them to potential advantages or disadvantages for users or stakeholders.

Now, is the following text describing content moderation as a problem, as a solution, or neither?

[Paste a tweet here and remove the brackets]

\subsubsection{Task 3: Policy Frames \newline}
“Content moderation” refers to the practice of screening and monitoring content posted by users on social media sites to determine if the content should be published or not, based on specific rules and guidelines.

I will ask you to classify a text as one of the frames defined below:

\begin{itemize}
    \item ECONOMY: The costs, benefits, or monetary/financial implications of the issue (to an individual, family, community, or to the economy as a whole).
    \item Capacity and resources: The lack of or availability of physical, geographical, spatial, human, and financial resources, or the capacity of existing systems and resources to implement or carry out policy goals.
    \item MORALITY: Any perspective—or policy objective or action (including proposed action)that is compelled by religious doctrine or interpretation, duty, honor, righteousness or any other sense of ethics or social responsibility.
    \item FAIRNESS AND EQUALITY: Equality or inequality with which laws, punishment, rewards, and resources are applied or distributed among individuals or groups. Also the balance between the rights or interests of one individual or group compared to another individual or group.
    \item POLICY PRESCRIPTION AND EVALUATION: Particular policies proposed for addressing an identified problem, and figuring out if certain policies will work, or if existing policies are effective.
    \item LAW AND ORDER, CRIME AND JUSTICE: Specific policies in practice and their enforcement, incentives, and implications. Includes stories about enforcement and interpretation of laws by individuals and law enforcement, breaking laws, loopholes, fines, sentencing and punishment. Increases or reductions in crime.
    \item SECURITY AND DEFENSE: Security, threats to security, and protection of one’s person, family, in-group, nation, etc. Generally an action or a call to action that can be taken to protect the welfare of a person, group, nation sometimes from a not yet manifested threat.
    \item HEALTH AND SAFETY: Health care access and effectiveness, illness, disease, sanitation, obesity, mental health effects, prevention of or perpetuation of gun violence, infrastructure and building safety.
    \item QUALITY OF LIFE: The effects of a policy on individuals’ wealth, mobility, access to resources, happiness, social structures, ease of day-to-day routines, quality of community life, etc.
    \item POLITICAL: Any political considerations surrounding an issue. Issue actions or efforts or stances that are political, such as partisan filibusters, lobbyist involvement, bipartisan efforts, deal-making and vote trading, appealing to one's base, mentions of political maneuvering. Explicit statements that a policy issue is good or bad for a particular political party.
    \item EXTERNAL REGULATION AND REPUTATION: The United States’ external relations with another nation; the external relations of one state with another; or relations between groups. This includes trade agreements and outcomes, comparisons of policy outcomes or desired policy outcomes.
    \item OTHER: Any topic that does not fit into the above categories.
\end{itemize}

Now, which of the above frames best fit the following text? Answer with only the option above that is most accurate and nothing else.

[Paste a tweet here and remove the brackets]

\subsubsection{Task 4: Stance Detection \newline}

“Content moderation” refers to the practice of screening and monitoring content posted by users on social media sites to determine if the content should be published or not, based on specific rules and guidelines. In the context of content moderation, Section 230 is a law in the United States that protects websites and other online platforms from being held legally responsible for the content posted by their users. This means that if someone posts something illegal or harmful on a website, the website itself cannot be sued for allowing it to be posted. However, websites can still choose to moderate content and remove anything that violates their own policies.

I will ask you to classify a text as in favor of, against, or neutral about Section 230:\\

A. “In favor of” expresses approval for Section 230 and/or advocates keeping Section 230\\
B. “Against” expresses disapproval towards Section 230 and/or advocates repealing Section 230\\
C. “Neutral” discusses Section 230 without expressing approval or disapproval towards it\\

Now, is the following text in favor of, against, or neutral about Section 230?

[Paste a tweet here and remove the brackets]

\subsubsection{Task 5: Topic Detection \newline}

“Content moderation” refers to the practice of screening and monitoring content posted by users on social media sites to determine if the content should be published or not, based on specific rules and guidelines.

I will ask you to classify a text as of the topics described below:

\begin{enumerate}
    \item Section 230, which is a law in the United States that protects websites and other online platforms from being held legally responsible for the content posted by their users (SECTION 230).
    \item The decision by many social media platforms, such as Twitter and Facebook, to suspend Donald Trump’s account (TRUMP BAN).
    \item Requests directed to Twitter’s support account or help center (TWITTER SUPPORT).
    \item Social media platforms’ policies and practices, such as community guidelines or terms of service (PLATFORM POLICIES).
    \item Complaints about platform’s policy and practices in deplatforming and content moderation or suggestions to suspend particular accounts, or complaints about accounts being suspended or reported  (COMPLAINTS).
    \item If a text is not about the SECTION 230, COMPLAINTS, TRUMP BAN, TWITTER SUPPORT, and PLATFORM POLICIES, then it should be classified in OTHER class (OTHER).
\end{enumerate}

Now, is the following text about SECTION 230, TRUMP BAN, COMPLAINTS, TWITTER SUPPORT, PLATFORM POLICIES, or OTHER?

[Paste a tweet here and remove the brackets]

\subsection{Dataset 2: Content Moderation Tweets (2023)}

\subsubsection{Task 1: Relevance \newline}

“Content moderation” refers to the practice of screening and monitoring content posted by users on social media sites to determine if the content should be published or not, based on specific rules and guidelines.

I will ask you to classify a text as relevant or irrelevant to the content moderation:

A: Text is RELEVANT if it includes: social media platforms’ content moderation rules and practices, censorship, governments’ regulation of online content moderation, and/or mild forms of content moderation like flagging, shadowbanning, or account suspension.

B: Text is IRRELEVANT if they do not refer to content moderation, as defined above. This would include, for example, a tweet by Trump that Twitter has labeled his tweet as “disputed”, or a tweet claiming that something is false.

Now, is the following text relevant or irrelevant to content moderation?

[Paste a tweet here and remove the brackets]

\subsubsection{Task 2: Problem/Solution Frames \newline}

“Content moderation” refers to the practice of screening and monitoring content posted by users on social media sites to determine if the content should be published or not, based on specific rules and guidelines.

I will ask you to classify a text as describing content moderation as a problem, as a solution, or neither:

A: Text describes content moderation as a PROBLEM if they emphasize negative effects of it, such as restrictions to free speech, censorship, or the biases that can emerge from decisions regarding what users are allowed to post.

B: Text describes content moderation as a SOLUTION if they emphasize positive effects of it, such as protecting users from harmful content such as hate speech, misinformation, illegal adult content, or spam. 

C: Text describes content moderation as NEUTRAL if they do not emphasize negative or positive effects of content moderation. For example if they simply report on the content moderation activity of social media platforms without linking them to potential advantages or disadvantages for users or stakeholders.

Now, is the following text describing content moderation as a problem, as a solution, or neither?

[Paste a tweet here and remove the brackets]

\subsection{Dataset 3: US Congress Members Tweets (2017-2022)}
\subsubsection{Task 1: Relevance \newline}
“Political content” refers to a text that pertains to politics or government policies at the local, national, or international level. This can include political figures, events, or issues, as well as text that uses political language or hashtags. 

I will ask you to classify a text as relevant or irrelevant to the political content:

Text is relevant if it uses political keywords or hashtags, mentions political figures or events, discusses policy issues such as immigration, abortion, foreign policy, health care, tax, or police shootings, or includes a link to news outlets or other political sources such as think tanks, political pundits or journalists, the White House, or the US Congress.
Text is irrelevant if it does not fit the criteria above

Now, is the following text relevant or irrelevant to political content?

[Paste a tweet here and remove the brackets]

\subsubsection{Task 2: Policy Frames \newline}
“Political content” refers to a text that pertains to politics or government policies at the local, national, or international level. This can include political figures, events, or issues, as well as text that uses political language or hashtags. 

I will ask you to classify a text as one of the frames defined below:

\begin{itemize}
    \item ECONOMY: The costs, benefits, or monetary/financial implications of the issue (to an individual, family, community, or to the economy as a whole).
    \item Capacity and resources: The lack of or availability of physical, geographical, spatial, human, and financial resources, or the capacity of existing systems and resources to implement or carry out policy goals.
    \item MORALITY: Any perspective—or policy objective or action (including proposed action)that is compelled by religious doctrine or interpretation, duty, honor, righteousness or any other sense of ethics or social responsibility.
    \item FAIRNESS AND EQUALITY: Equality or inequality with which laws, punishment, rewards, and resources are applied or distributed among individuals or groups. Also the balance between the rights or interests of one individual or group compared to another individual or group.
    \item POLICY PRESCRIPTION AND EVALUATION: Particular policies proposed for addressing an identified problem, and figuring out if certain policies will work, or if existing policies are effective.
    \item LAW AND ORDER, CRIME AND JUSTICE: Specific policies in practice and their enforcement, incentives, and implications. Includes stories about enforcement and interpretation of laws by individuals and law enforcement, breaking laws, loopholes, fines, sentencing and punishment. Increases or reductions in crime.
    \item SECURITY AND DEFENSE: Security, threats to security, and protection of one’s person, family, in-group, nation, etc. Generally an action or a call to action that can be taken to protect the welfare of a person, group, nation sometimes from a not yet manifested threat.
    \item HEALTH AND SAFETY: Health care access and effectiveness, illness, disease, sanitation, obesity, mental health effects, prevention of or perpetuation of gun violence, infrastructure and building safety.
    \item QUALITY OF LIFE: The effects of a policy on individuals’ wealth, mobility, access to resources, happiness, social structures, ease of day-to-day routines, quality of community life, etc.
    \item POLITICAL: Any political considerations surrounding an issue. Issue actions or efforts or stances that are political, such as partisan filibusters, lobbyist involvement, bipartisan efforts, deal-making and vote trading, appealing to one's base, mentions of political maneuvering. Explicit statements that a policy issue is good or bad for a particular political party.
    \item EXTERNAL REGULATION AND REPUTATION: The United States’ external relations with another nation; the external relations of one state with another; or relations between groups. This includes trade agreements and outcomes, comparisons of policy outcomes or desired policy outcomes.
    \item OTHER: Any topic that does not fit into the above categories.
\end{itemize}

Now, which of the above frames best fit the following text? Answer with only the option above that is most accurate and nothing else.

[Paste a tweet here and remove the brackets]

\subsection{Dataset 4: Content Moderation News Articles (2020-2021)}
\subsubsection{Task 1: Relevance \newline}

“Content moderation” refers to the practice of screening and monitoring content posted by users on social media sites to determine if the content should be published or not, based on specific rules and guidelines.

I will ask you to classify a text as relevant or irrelevant to the content moderation:

A: Text is RELEVANT if it includes: social media platforms’ content moderation rules and practices, censorship, governments’ regulation of online content moderation, and/or mild forms of content moderation like flagging, shadowbanning, or account suspension.

B: Text is IRRELEVANT if they do not refer to content moderation, as defined above. This would include, for example, a tweet by Trump that Twitter has labeled his tweet as “disputed”, or a tweet claiming that something is false.

Now, is the following text relevant or irrelevant to content moderation?

[Paste a news article text here and remove the brackets]

\subsubsection{Task 2: Problem/Solution Frames \newline}

“Content moderation” refers to the practice of screening and monitoring content posted by users on social media sites to determine if the content should be published or not, based on specific rules and guidelines.

I will ask you to classify a text as describing content moderation as a problem, as a solution, or neither:

A: Text describes content moderation as a PROBLEM if they emphasize negative effects of it, such as restrictions to free speech, censorship, or the biases that can emerge from decisions regarding what users are allowed to post.

B: Text describes content moderation as a SOLUTION if they emphasize positive effects of it, such as protecting users from harmful content such as hate speech, misinformation, illegal adult content, or spam. 

C: Text describes content moderation as NEUTRAL if they do not emphasize negative or positive effects of content moderation. For example if they simply report on the content moderation activity of social media platforms without linking them to potential advantages or disadvantages for users or stakeholders.

Now, is the following text describing content moderation as a problem, as a solution, or neither?

[Paste a news article text here and remove the brackets]

\section{Few-Shot Annotation Codebook (Chain-of-Thought)} \label{appendix:codebook_fewshot_cot}

\subsection{Dataset 1: Content Moderation Tweets (2020-2021)}
\subsubsection{Task 1: Relevance \newline}

“Content moderation” refers to the practice of screening and monitoring content posted by users on social media sites to determine if the content should be published or not, based on specific rules and guidelines.

I will ask you to classify a text as relevant or irrelevant to the content moderation:

A: Text is RELEVANT if it includes: social media platforms’ content moderation rules and practices, censorship, governments’ regulation of online content moderation, and/or mild forms of content moderation like flagging, shadowbanning, or account suspension.

B: Text is IRRELEVANT if they do not refer to content moderation, as defined above. This would include, for example, a tweet by Trump that Twitter has labeled his tweet as “disputed”, or a tweet claiming that something is false.

The tweets below are some examples of 'RELEVANT' or 'IRRELEVANT' texts to content moderation in the form of question and answer: 

Question: Please stop lying.cWe all know you can’t stand Sanders and the platform he stands for and people that support progressive policies.cBut can you at least not LIE? It’s a really bad look. Thanks.

Answer: Based on the given criteria, the text provided is IRRELEVANT to content moderation. It doesn't discuss social media platforms' content moderation rules and practices, censorship, governments' regulation of online content moderation, or any forms of content moderation like flagging, shadowbanning, or account suspension. Instead, it appears to be a message directed at individuals and their opinions about Sanders and progressive policies.

Question: Hey guys, my TL is hella cluttered from all the follow backs I did this season. I’m going to reinstate my “if you only rt content” I’m not going to follow you policy. No hard feelings and I hope you stick around, but if I unfollow you that’s likely why. 

Answer: Based on the given criteria, the text provided is IRRELEVANT to content moderation. It doesn't discuss social media platforms' content moderation rules and practices, censorship, governments' regulation of online content moderation, or any mild forms of content moderation like flagging, shadowbanning, or account suspension. Instead, it appears to be a personal statement about cluttered timelines and a follow-back policy on Twitter.

Question: New from Pediatrics: social media influencers generate millions of impressions for unhealthy food and drink brands through product placement. The authors call on the FTC to strengthen regulations on YouTube product placement featuring young children. \url{https://t.co/Yicupispun}

Answer: Based on the given criteria, the text provided is RELEVANT to content moderation. It discusses the impact of social media influencers and product placement on unhealthy food and drink brands. It also mentions calling on the FTC (Federal Trade Commission) to strengthen regulations on YouTube product placement featuring young children. This relates to content moderation as it involves discussing regulations and practices related to content promotion and advertising on social media platforms.

Question: This is a tweet from today of the sitting President of the United States accusing a respected journalist and outspoken critic of his of MURDER with NO EVIDENCE whatsoever. Zero. Twitter should immediately remove this tweet and suspend this account. This is INSANE. \url{https://t.co/chlp7iKBXL}

Answer: Based on the given criteria, the text provided is RELEVANT to content moderation. It discusses the actions and potential response of Twitter in relation to a tweet made by the sitting President of the United States. It addresses the need for Twitter to remove the tweet and suspend the account due to the serious accusation made without evidence. This involves discussing content moderation rules and practices, specifically regarding the removal of content and potential account suspension.

Now, is the following text relevant or irrelevant to content moderation?

[Paste a tweet here and remove the brackets]

\subsubsection{Task2: Problem/Solution Frames \newline}

“Content moderation” refers to the practice of screening and monitoring content posted by users on social media sites to determine if the content should be published or not, based on specific rules and guidelines.

I will ask you to classify a text as describing content moderation as a problem, as a solution, or neither:

A: Text describes content moderation as a PROBLEM if they emphasize negative effects of it, such as restrictions to free speech, censorship, or the biases that can emerge from decisions regarding what users are allowed to post.

B: Text describes content moderation as a SOLUTION if they emphasize positive effects of it, such as protecting users from harmful content such as hate speech, misinformation, illegal adult content, or spam. 

C: Text describes content moderation as NEUTRAL if they do not emphasize negative or positive effects of content moderation. For example if they simply report on the content moderation activity of social media platforms without linking them to potential advantages or disadvantages for users or stakeholders.

The following tweets are some examples of 'PROBLEM', 'SOLUTION' or 'NEUTRAL' texts about content moderation in the form of question and answer:

Question: \@amerxmirza I have reported him to make sure his account gets suspended. Disgusting and derogation remarks shouldn't be tolerated. Please report. Thanks

Answer: The given text can be classified as describing content moderation as a SOLUTION. It emphasizes the positive effect of content moderation by advocating for reporting and suspending the account of a user who made "disgusting and derogatory remarks." The text acknowledges the need to not tolerate such behavior and encourages others to report it, highlighting the protective role of content moderation in addressing harmful content.

Question: Channel link  \url{https://t.co/9ms7ukCI84}  please block this channel this content very harmful \url{https://t.co/zeJ512JPPx}

Answer: The given text can be classified as describing content moderation as a SOLUTION. It emphasizes the positive effect of content moderation by requesting the blocking of a specific channel due to its harmful content. The text recognizes the need to take action and protect users from potentially harmful content, aligning with the protective role of content moderation in addressing harmful material.

Question: Finally could restore my account. Why twitter even require phone number in the first place? Aren't we supposed to have a choice not to show it thus not being banned after, if twitter bots decide you're one of them? \#twitter \#ElonMusk \#FreedomInternet

Answer: The given text can be classified as describing content moderation as a PROBLEM. It criticizes the requirement of providing a phone number by Twitter and questions the lack of choice in not showing it. The text implies frustration with the potential consequences of content moderation decisions made by Twitter bots, suggesting concerns about restrictions to freedom and potential biases in the platform's moderation practices.

Question: Of all the issues we’re facing now, why’s social media regulation the goal for this people? Freedom of speech is a basic human right and any attempt to take it from Nigerians should be rejected \#RejectNITDAbill \#Notosocialmediaregulation

Answer: The given text can be classified as describing content moderation as a PROBLEM. It emphasizes the negative effects of social media regulation, specifically mentioning the potential infringement on freedom of speech as a basic human right. The text opposes any attempt to take away this right from Nigerians, suggesting concerns about restrictions and censorship associated with content moderation.

Question: \@BotGerad \@R6Jena \@zxdd account suspended xd

Answer: Based on the given text, it does not explicitly emphasize the negative or positive effects of content moderation. Instead, it simply reports the suspension of an account without further elaboration. Therefore, the text can be classified as describing content moderation as NEUTRAL, as it does not emphasize either the problems or the solutions associated with content moderation.

Question: \@elonmusk \@ellagirwin  \@esthercrawford there seems to have been some mistake, renowned internet love-monger \@nickjfuentes was suspended from Twitter. Please do the right thing and re-instate his account for the sake of humanity

Answer: Based on the given text, it is possible to interpret it differently. While the text does request the reinstatement of a suspended account, it does not explicitly mention any negative effects or problems related to content moderation. Therefore, an alternative classification could be that the text describes content moderation as NEUTRAL since it does not emphasize negative or positive effects. It simply requests the reinstatement of a specific account without further elaboration on the broader implications of content moderation.

Now, is the following text describing content moderation as a problem, as a solution, or neither?

[Paste a tweet here and remove the brackets]

\subsubsection{Task 3: Policy Frames \newline}
“Content moderation” refers to the practice of screening and monitoring content posted by users on social media sites to determine if the content should be published or not, based on specific rules and guidelines.

I will ask you to classify a text as one of the frames defined below:

\begin{itemize}
    \item ECONOMY: The costs, benefits, or monetary/financial implications of the issue (to an individual, family, community, or to the economy as a whole).
    \item Capacity and resources: The lack of or availability of physical, geographical, spatial, human, and financial resources, or the capacity of existing systems and resources to implement or carry out policy goals.
    \item MORALITY: Any perspective—or policy objective or action (including proposed action)that is compelled by religious doctrine or interpretation, duty, honor, righteousness or any other sense of ethics or social responsibility.
    \item FAIRNESS AND EQUALITY: Equality or inequality with which laws, punishment, rewards, and resources are applied or distributed among individuals or groups. Also the balance between the rights or interests of one individual or group compared to another individual or group.
    \item POLICY PRESCRIPTION AND EVALUATION: Particular policies proposed for addressing an identified problem, and figuring out if certain policies will work, or if existing policies are effective.
    \item LAW AND ORDER, CRIME AND JUSTICE: Specific policies in practice and their enforcement, incentives, and implications. Includes stories about enforcement and interpretation of laws by individuals and law enforcement, breaking laws, loopholes, fines, sentencing and punishment. Increases or reductions in crime.
    \item SECURITY AND DEFENSE: Security, threats to security, and protection of one’s person, family, in-group, nation, etc. Generally an action or a call to action that can be taken to protect the welfare of a person, group, nation sometimes from a not yet manifested threat.
    \item HEALTH AND SAFETY: Health care access and effectiveness, illness, disease, sanitation, obesity, mental health effects, prevention of or perpetuation of gun violence, infrastructure and building safety.
    \item QUALITY OF LIFE: The effects of a policy on individuals’ wealth, mobility, access to resources, happiness, social structures, ease of day-to-day routines, quality of community life, etc.
    \item POLITICAL: Any political considerations surrounding an issue. Issue actions or efforts or stances that are political, such as partisan filibusters, lobbyist involvement, bipartisan efforts, deal-making and vote trading, appealing to one's base, mentions of political maneuvering. Explicit statements that a policy issue is good or bad for a particular political party.
    \item EXTERNAL REGULATION AND REPUTATION: The United States’ external relations with another nation; the external relations of one state with another; or relations between groups. This includes trade agreements and outcomes, comparisons of policy outcomes or desired policy outcomes.
    \item OTHER: Any topic that does not fit into the above categories.
\end{itemize}

The following tweets are some examples of these frames in the form of question and answer:

Question: TY \@AGSchneiderman for your investigation into red light camera violations by school bus companies. Our children’s safety is paramount. \url{https://t.co/CVLj7Lk6r4}

Answer: HEALTH AND SAFETY

Question: Had a huge turnout for the \#GrowWithGoogle digital skills event, which I hosted with \@google today in Charleston! Programs like this one will help strengthen our workforce \& create a brighter economic future for West Virginians. \url{https://t.co/3ma7Mv1EBR}

Answer: ECONOMY

Question: On the floor about to offer and speak on my amendment to eliminate subsistence fees at BOP halfway houses, which house DC Code felons.

Answer: Based on the provided text, the most accurate frame would be: POLICY PRESCRIPTION AND EVALUATION. The text mentions the speaker's intention to offer an amendment to eliminate subsistence fees at BOP halfway houses that house DC Code felons. This indicates a specific policy proposal aimed at addressing a problem related to subsistence fees for felons in halfway houses. Therefore, it falls under the frame of policy prescription and evaluation.

Question: It is refreshing that the \#JointAddress tonight was a call to unite the country around a strategy for a stronger, brighter future.

Answer: Based on the provided text, the most accurate frame would be: POLITICAL. The text refers to the \#JointAddress, which typically refers to a political event where the President of the United States addresses the nation. It mentions the call to unite the country around a strategy for a stronger, brighter future, which indicates a political consideration aimed at rallying support and fostering unity. Therefore, the text falls under the frame of political considerations.

Question: Today we remember and honor the men and women of our Armed Forces who remain missing in action or prisoners of war. \#POWMIARecognitionDay \url{https://t.co/D9z1akkjKW}

Answer: The text you provided can be classified under the frame of SECURITY AND DEFENSE. It is a statement that acknowledges and honors the men and women of the Armed Forces who are missing in action or prisoners of war. It emphasizes the importance of remembering and honoring their sacrifice, which relates to the protection and welfare of individuals and the nation's security.

Question: No longer can POTUS pretend that Putin’s Russia was not responsible for cyberattacks targeting the 2016 election. Today’s indictments are another example of why the Mueller investigation must continue unimpeded. The American people need to learn the truth. \url{https://t.co/mYwE4p4jR4}

Answer: The following frame is the best fit: EXTERNAL REGULATION AND REPUTATION: The text highlights the responsibility of Putin's Russia for cyberattacks targeting the 2016 election and emphasizes the need for the Mueller investigation to continue unimpeded. It mentions the truth that the American people need to learn. This relates to the external relations and reputation of the United States and its interactions with another nation (Russia) in the context of cyberattacks and election interference.

Question: Women have a valuable place in STEAM fields—let’s make sure they have a seat at the table. \url{https://t.co/LhOawvSszP}

Answer: Based on the provided text, the most accurate classification would be: FAIRNESS AND EQUALITY. The text emphasizes the importance of ensuring women have a seat at the table in STEAM (Science, Technology, Engineering, Arts, and Mathematics) fields. It addresses the equality or inequality of opportunities and representation for women in these fields, highlighting the need for fairness and equal participation.

Question: BREAKING: \@HouseIntelComm Chairman Conaway just announced that the Committee is closing the "Russian collusion" investigation, having found zero evidence of any collusion between the Trump campaign and Russians. Case closed. It's time we return focus to the people's agenda.

Answer: The text can be classified under the following frame: LAW AND ORDER, CRIME AND JUSTICE. The text mentions the investigation into "Russian collusion" and the finding of zero evidence of collusion between the Trump campaign and Russians. It highlights the closure of the investigation and suggests a shift in focus to the people's agenda. This classification relates to specific policies in practice (investigation), enforcement (finding of evidence), and implications (closure) within the realm of law and order, crime, and justice.

Question: The Trump-Sessions "zero tolerance" family separation border policies are not required, right or moral. \url{https://t.co/aAFX8Q6eKT}

Answer: Based on the provided text, the best fit among the defined frames would be: MORALITY. The text expresses a perspective on the Trump-Sessions "zero tolerance" family separation border policies, stating that they are not required, right, or moral. This indicates a moral evaluation of the policies based on ethics and social responsibility.

Question: Glad to work with \@TomRooney \@RepSeanMaloney \@RepStefanik on this bipartisan bill to help our homeless veterans. \#HousingOurHeroes

Answer: The text highlights a bipartisan bill aimed at helping homeless veterans, which directly relates to improving the quality of life for these individuals. The focus is on the effects of the policy on individuals' well-being, access to resources, and housing stability.
Therefore, the frame "QUALITY OF LIFE" is the most accurate classification for the given text.

Question: Prayers to my dear friend \@SteveScalise \&all injured after this morning's horrific shooting.Thinking of their families in this shocking time

Answer: The text expresses sympathy and concern for the injured individuals and their families after a shooting incident, but it doesn't explicitly address the broader implications or aspects related to economy, morality, fairness and equality, policy prescription and evaluation, law and order, crime and justice, health and safety, quality of life, political considerations, external regulation and reputation, or any other specific frame. Therefore, the most accurate classification for this text would be "OTHER," as it doesn't fit into the predefined frames provided.

Now, which of the above frames best fit the following text? Answer with only the option above that is most accurate and nothing else.

[Paste a tweet here and remove the brackets]

\subsubsection{Task 4: Stance Detection \newline}

“Content moderation” refers to the practice of screening and monitoring content posted by users on social media sites to determine if the content should be published or not, based on specific rules and guidelines. In the context of content moderation, Section 230 is a law in the United States that protects websites and other online platforms from being held legally responsible for the content posted by their users. This means that if someone posts something illegal or harmful on a website, the website itself cannot be sued for allowing it to be posted. However, websites can still choose to moderate content and remove anything that violates their own policies.

I will ask you to classify a text as in favor of, against, or neutral about Section 230:\\

A. “In favor of” expresses approval for Section 230 and/or advocates keeping Section 230\\
B. “Against” expresses disapproval towards Section 230 and/or advocates repealing Section 230\\
C. “Neutral” discusses Section 230 without expressing approval or disapproval towards it\\

The following tweets are some examples of texts “IN FAVOR OF” ,“AGAINST” or “NEUTRAL” about section230 in the form of question and answer:

Question: \#Section230 is poorly understood by lawmakers on both sides of the aisle, and it is dangerous for them to use it as a political football. To save online free speech, we must \#Protect230 Contact lawmakers: \url{https://t.co/ldSL75knH4}

Answer: The provided text is in favor of Section 230 and advocates keeping it. The use of the hashtag \#Protect230 and the statement "To save online free speech, we must \#Protect230" indicate support for the law and its role in preserving online free speech. Therefore, the text can be classified as "IN FAVOR OF" Section 230.

Question: Facebook and Twitter CEOs warn against demolishing Section 230, the law that shields tech giants https://t.co/CItuLmTTxE by \@alexiskweed \url{https://t.co/7Y6eG19YoZ}

Answer: The provided text is in favor of Section 230. It mentions that the CEOs of Facebook and Twitter are warning against demolishing Section 230, indicating their support for the law. Therefore, the text can be classified as "IN FAVOR OF" Section 230.

Question: How do we get big tech companies like Twitter to abide by the spirit of the 1st Amendment, or moderate content by a set of clear and definable standards? Canyon Brimhall joins in the third episode of our series on big tech, free speech, and Section 230. \url{https://t.co/RfoJVuQPEh}

Answer: The provided text discusses the need for big tech companies to abide by the spirit of the First Amendment and moderate content according to clear and definable standards. It mentions Section 230 in the context of a discussion about big tech, free speech, and content moderation. The text does not express explicit approval or disapproval towards Section 230 itself but raises the question of how to ensure big tech companies adhere to the principles of the First Amendment. Therefore, the text can be classified as "NEUTRAL" about Section 230.

Question: I sat down with the Meme King himself \@bennyjohnson in Washington, DC to discuss Big Tech, Social Media Censorship and its impact on the election, Section 230, 
PLEASE DO NOT MISS THIS! \url{https://t.co/ETWKY16J8d}

Answer: The provided text does not clearly express whether it is in favor of or against Section 230. It discusses various topics such as Big Tech, social media censorship, Section 230, and its impact on the election without explicitly stating approval or disapproval towards Section 230. Therefore, the text can be classified as "NEUTRAL" about Section 230.

Question: \@JoeBiden Trump MUST be re-elected. We MUST vote in GOP in the House, keep the Senate and we MUST eliminate Section 230 protections for big tech social media! Social media \& Corporate America will kill our free speech! BLM is a Marxist group looking to DESTROY USA! \url{https://t.co/R37Qi2rlNy}

Answer: The provided text is against Section 230. It advocates for eliminating Section 230 protections for big tech social media and expresses concerns about the impact of social media and corporate America on free speech. Therefore, the text can be classified as "AGAINST" Section 230.

Question: BREAKING: Supreme Court Justice Clarence Thomas finds social media companies do not have First Amendment right to ban protected speech and that Section 230 is unconstitutional

Answer: The provided text expresses a critical stance towards Section 230 by stating that Supreme Court Justice Clarence Thomas finds it unconstitutional and that social media companies do not have a First Amendment right to ban protected speech. Therefore, the text can be classified as "AGAINST" Section 230.

Now, is the following text in favor of, against, or neutral about Section 230?

[Paste a tweet here and remove the brackets]

\subsubsection{Task 5: Topic Detection \newline}

“Content moderation” refers to the practice of screening and monitoring content posted by users on social media sites to determine if the content should be published or not, based on specific rules and guidelines.

I will ask you to classify a text as of the topics described below:

\begin{enumerate}
    \item Section 230, which is a law in the United States that protects websites and other online platforms from being held legally responsible for the content posted by their users (SECTION 230).
    \item The decision by many social media platforms, such as Twitter and Facebook, to suspend Donald Trump’s account (TRUMP BAN).
    \item Requests directed to Twitter’s support account or help center (TWITTER SUPPORT).
    \item Social media platforms’ policies and practices, such as community guidelines or terms of service (PLATFORM POLICIES).
    \item Complaints about platform’s policy and practices in deplatforming and content moderation or suggestions to suspend particular accounts, or complaints about accounts being suspended or reported  (COMPLAINTS).
    \item If a text is not about the SECTION 230, COMPLAINTS, TRUMP BAN, TWITTER SUPPORT, and PLATFORM POLICIES, then it should be classified in OTHER class (OTHER).
\end{enumerate}

The following tweets are some examples of the topics described above in the form of question and answer:

Question: \@TangledUpInDead \@soupmaned \@jkosseff \@jenniferm\_q The referenced case has absolutely nothing to do with Section 230.

Answer: Based on the provided text, it falls under the "SECTION 230" category. The text explicitly mentions Section 230 and states that the referenced case is unrelated to it.

Question: \@LindseyGrahamSC The fact that you call it a “demand” is abhorrent! The American people deserve the support of their government and \$2,000 pp is well overdue. The fact it will be held hostage unless Section 230 is addressed is horseshit! \#humanroulette \#americansaredyingtrumpkeepslying

Answer: Based on the provided text, it falls under the "SECTION 230" category. The text mentions Section 230 and expresses frustration about it being used as leverage to address other issues, specifically relating to the demand for \$2,000 per person.

Question: \@YourAnonCentral Hack Twitter and Ban Trump

Answer: Based on the given text, it would be classified as a statement related to "TRUMP BAN," specifically mentioning hacking Twitter and advocating for the banning of Trump.

Question: Russian-made \#Telegram messenger shoots to top of US app charts, amid fears of wider social media crackdown following Trump ban — RT Russia \& Former Soviet Union \url{https://t.co/3HGLCNNJ2T}

Answer: Based on the given text, it would be classified as a statement related to "TRUMP BAN." The text discusses fears of a wider social media crackdown following Trump's ban and mentions the rise of the Russian-made Telegram messenger.

Question: Twitter decided to ban Mike Lindell, who founded bedding company My Pillow, due to “repeated violations” of its civic integrity policy, a spokesperson said in a statement. The policy was implemented last September and is targeted at fighting disinformation \url{https://t.co/lM2FeHuv3f}

Answer: This tweet appears to be more closely related to "PLATFORM POLICIES". The text discusses Twitter's decision to ban Mike Lindell based on violations of its civic integrity policy, which falls under the category of social media platforms' policies and practices.

Question: Twitter To Introduce This New Feature To Curb The Spread Of Fake, Harmful Content. \url{https://t.co/TqHeINqeQzhttps://t.co/SrwaCC5fdL} So many \#cherries to choose from this season! Choose your favourite or get them all on \url{https://t.co/gAhHib40nQ}. CherrySeason \#FreshFruits \#superplum \#FarmFresh \#VocalForLocal \#StellaCherries \#OrderOnline \#HomeDelivery \#BlackCherries \#MerchantCherries https://t.co/WNpIDh72p3

Answer: This tweet should be classified as "PLATFORM POLICIES". The text mentions Twitter's introduction of a new feature to curb the spread of fake and harmful content. This aspect relates to social media platform policies aimed at content moderation.

Question: This post has been up 24 minutes and has 12 impressions \@Twitter \@TwitterSupport. I have 3800 followers. This number is about 95\% off where it should be Why are you secretly shadow banning me? \url{https://t.co/l1oF7lqraJ}

Answer: The text is about TWITTER SUPPORT. The author is addressing Twitter and TwitterSupport and complaining about a decrease in his/her number of followers, suggesting that he/she is being shadowbanned by Twitter. The use of the words "impressions" and "followers" suggests that the author is concerned with the platform's policies and practices regarding content moderation and deplatforming.

Question: Using a criminal's picture insulting our history and culture this is where I should report this user to \@Twitter \@TwitterSupport for hateful content  and threatening \#CyberSecurity \url{https://t.co/KdIinpgMXf}

Answer: The tweet is about TWITTER SUPPORT because it mentions reporting a user to \@Twitter and \@TwitterSupport for hateful content and threatening \#CyberSecurity. This indicates a request for assistance from Twitter's support team to moderate and remove potentially harmful content shared on the platform.

Question: You may have agreed or disagreed with JD from NY, but I do not think anyone should celebrate deplatforming

Answer: This text falls under the "COMPLAINTS" class as it expresses dissatisfaction with deplatforming and suggests that it should not be celebrated, but it does not refer specifically to any social media platform, policy, or practice.

Question: \@dbongino I found out Twitter is shadowbanning me and I'm essentially a nobody.  How many other people with viewpoints they do not like and being placed behind a wall?

Answer: The given text is about COMPLAINTS. The text talks about an individual named "\@dbongino" who has found out that Twitter has shadowbanned them, resulting in their tweets being placed behind a wall. The user expresses concern about how many other people with differing viewpoints might also be affected by Twitter's policies. This text is a complaint about Twitter's content moderation and deplatforming practices, and hence it falls under the COMPLAINTS category.

Question: \@propjoesays Twitter needs to issue a blanket ban on this kind of fearmongering. Anyone engaging in voter suppression or questioning the integrity of elections needs to have their account suspended.

Answer: The text is not about the topics mentioned in the instructions, Instead, it is focused on a specific suggestion that Twitter should issue a blanket ban on certain types of content related to voter suppression and questioning the integrity of elections. As it does not fit into any of the given topic categories, it should be classified under OTHER.

Question: Lots of people have been noticing that recently. They are big time shadow banning. It’s so obvious. There are people I follow who tweet a couple times and hour and thought they were off twitter because I haven’t seen a tweet in months \url{https://t.co/IQkZ2346VB}

Answer: The tweet is more appropriately classified as "OTHER." It does not specifically mention Section 230, the Trump ban, Twitter support, or platform policies. Instead, it discusses shadow banning and the author's observation of decreased visibility of tweets from certain accounts. Since it does not align with any of the specified topics, "OTHER" is the most suitable classification.

Now, is the following text about SECTION 230, TRUMP BAN, COMPLAINTS, TWITTER SUPPORT, PLATFORM POLICIES, or OTHER?

[Paste a tweet here and remove the brackets]

\subsection{Dataset 2: Content Moderation Tweets (2023)}

\subsubsection{Task 1: Relevance \newline}

“Content moderation” refers to the practice of screening and monitoring content posted by users on social media sites to determine if the content should be published or not, based on specific rules and guidelines.

I will ask you to classify a text as relevant or irrelevant to the content moderation:

A: Text is RELEVANT if it includes: social media platforms’ content moderation rules and practices, censorship, governments’ regulation of online content moderation, and/or mild forms of content moderation like flagging, shadowbanning, or account suspension.

B: Text is IRRELEVANT if they do not refer to content moderation, as defined above. This would include, for example, a tweet by Trump that Twitter has labeled his tweet as “disputed”, or a tweet claiming that something is false.

The tweets below are some examples of 'RELEVANT' and 'IRRELEVANT' texts to content moderation in the form of question and answer: 

Question: \@GillianBateman5 \@ankertilbrewer \@BreesAnna She blocked me on my old account but added me on this one for some reason

Answer: Based on the criteria provided, the text provided is IRRELEVANT to content moderation. It does not discuss social media platforms' content moderation rules and practices, censorship, governments' regulation of online content moderation, or any forms of content moderation such as flagging, shadowbanning, or account suspension. Instead, it appears to be a casual conversation or observation about someone blocking and adding people on social media.

Question: This is how will \#SriLanka debt restructuring look like? accordingly to VERITÉ RESEARCH SRI LANKA POLICY GROUP \@VeriteResearch \@CBSL \@StanChart \url{https://t.co/adjf5cKzxnhttps://t.co/Er1nP9a4jh}

Answer: Based on the given criteria, the text provided is IRRELEVANT to content moderation. It does not discuss social media platforms' content moderation rules and practices, censorship, governments' regulation of online content moderation, or any mild forms of content moderation. Instead, the text appears to be about the debt restructuring in Sri Lanka, which is unrelated to content moderation on social media sites.

Question Dear \@Twitter \@TwitterMENA \@TwitterSupport \@SalmaMMMT account has been suspended 1 day ago And without any reason

Answer: Based on the given criteria, the text provided is RELEVANT to content moderation. It mentions the suspension of an account on Twitter (\@SalmaMMMT) without any given reason. This relates to content moderation as it touches upon the action of account suspension, which is a mild form of content moderation employed by social media platforms.

Question: Finally could restore my account. Why twitter even require phone number in the first place? Aren't we supposed to have a choice not to show it thus not being banned after, if twitter bots decide you're one of them? \#twitter \#ElonMusk \#FreedomInternet

Answer: The text provided can be considered RELEVANT to content moderation. Although it primarily discusses personal account issues and frustrations with Twitter's phone number requirement, it also raises concerns about being banned based on the decisions of Twitter bots. This alludes to the practice of content moderation, where automated systems are often involved in flagging or suspending accounts based on certain criteria.

Now, is the following text relevant or irrelevant to content moderation?

[Paste a tweet here and remove the brackets]

\subsubsection{Task 2: Problem/Solution Frames \newline}

“Content moderation” refers to the practice of screening and monitoring content posted by users on social media sites to determine if the content should be published or not, based on specific rules and guidelines.

I will ask you to classify a text as describing content moderation as a problem, as a solution, or neither:

A: Text describes content moderation as a PROBLEM if they emphasize negative effects of it, such as restrictions to free speech, censorship, or the biases that can emerge from decisions regarding what users are allowed to post.

B: Text describes content moderation as a SOLUTION if they emphasize positive effects of it, such as protecting users from harmful content such as hate speech, misinformation, illegal adult content, or spam. 

C: Text describes content moderation as NEUTRAL if they do not emphasize negative or positive effects of content moderation. For example if they simply report on the content moderation activity of social media platforms without linking them to potential advantages or disadvantages for users or stakeholders.

The following texts are some examples of 'PROBLEM', 'SOLUTION' or 'NEUTRAL' texts about content moderation in the form of question and answer:

Question: \@amerxmirza I have reported him to make sure his account gets suspended. Disgusting and derogation remarks shouldn't be tolerated. Please report. Thanks

Answer: The given text can be classified as describing content moderation as a SOLUTION. It emphasizes the positive effect of content moderation by advocating for reporting and suspending the account of a user who made "disgusting and derogatory remarks." The text acknowledges the need to not tolerate such behavior and encourages others to report it, highlighting the protective role of content moderation in addressing harmful content.

Question: Channel link \url{https://t.co/9ms7ukCI84} please block this channel this content very harmful \url{https://t.co/zeJ512JPPx}

Answer: The given text can be classified as describing content moderation as a SOLUTION. It emphasizes the positive effect of content moderation by requesting the blocking of a specific channel due to its harmful content. The text recognizes the need to take action and protect users from potentially harmful content, aligning with the protective role of content moderation in addressing harmful material.

Question: Finally could restore my account. Why twitter even require phone number in the first place? Aren't we supposed to have a choice not to show it thus not being banned after, if twitter bots decide you're one of them? 
\#twitter \#ElonMusk \#FreedomInternet

Answer: The given text can be classified as describing content moderation as a PROBLEM. It criticizes the requirement of providing a phone number by Twitter and questions the lack of choice in not showing it. The text implies frustration with the potential consequences of content moderation decisions made by Twitter bots, suggesting concerns about restrictions to freedom and potential biases in the platform's moderation practices.

Question: Of all the issues we’re facing now, why’s social media regulation the goal for this people? Freedom of speech is a basic human right and any attempt to take it from Nigerians should be rejected \#RejectNITDAbill \#Notosocialmediaregulation

Answer: The given text can be classified as describing content moderation as a PROBLEM. It emphasizes the negative effects of social media regulation, specifically mentioning the potential infringement on freedom of speech as a basic human right. The text opposes any attempt to take away this right from Nigerians, suggesting concerns about restrictions and censorship associated with content moderation.

Question: \@BotGerad \@R6Jena \@zxdd account suspended xd

Answer: Based on the given text, it does not explicitly emphasize the negative or positive effects of content moderation. Instead, it simply reports the suspension of an account without further elaboration. Therefore, the text can be classified as describing content moderation as NEUTRAL, as it does not emphasize either the problems or the solutions associated with content moderation.

Question: \@elonmusk \@ellagirwin  \@esthercrawford there seems to have been some mistake, renowned internet love-monger \@nickjfuentes was suspended from Twitter. Please do the right thing and re-instate his account for the sake of humanity 

Answer: Based on the given text, it is possible to interpret it differently. While the text does request the reinstatement of a suspended account, it does not explicitly mention any negative effects or problems related to content moderation. Therefore, an alternative classification could be that the text describes content moderation as NEUTRAL since it does not emphasize negative or positive effects. It simply requests the reinstatement of a specific account without further elaboration on the broader implications of content moderation.

Now, is the following text describing content moderation as a problem, as a solution, or neither?

[Paste a tweet here and remove the brackets]

\subsection{Dataset 3: US Congress Members Tweets (2017-2022)}
\subsubsection{Task 1: Relevance \newline}
“Political content” refers to a text that pertains to politics or government policies at the local, national, or international level. This can include political figures, events, or issues, as well as text that uses political language or hashtags. 

I will ask you to classify a text as relevant or irrelevant to the political content:

Text is relevant if it uses political keywords or hashtags, mentions political figures or events, discusses policy issues such as immigration, abortion, foreign policy, health care, tax, or police shootings, or includes a link to news outlets or other political sources such as think tanks, political pundits or journalists, the White House, or the US Congress.
Text is irrelevant if it does not fit the criteria above

The following tweets are some examples of RELEVANT or IRRELEVANT texts to political contents in the form of question and answer:

Question: What inspires me: being a warrior for human dignity, human potential. \url{https://t.co/k6NXxcThaD}

Answer: Based on the given criteria, the text you provided is IRRELEVANT to political content. It does not include any political keywords or hashtags, mention political figures or events, discuss policy issues, or provide links to political sources. Instead, it focuses on concepts such as human dignity and human potential.

Question: I hope this time brings you the opportunity to reflect, to \#hope, and to know that you're loved by God. \#MerryChristmas, \#GA09! \url{https://t.co/cOfwb9TxWa}

Answer: Based on the provided criteria, the text you provided is IRRELEVANT to political content. While it includes hashtags such as \#hope, \#MerryChristmas, and \#GA09, it does not use any political keywords or discuss political figures, events, or policy issues. The focus of the text is on reflection, hope, and sending Christmas wishes.

Question: Pres Trump wants another 1 percent for the richest 1 percent. 100 percent irresponsible. \url{https://t.co/CawU4SpUTo}

Answer: Based on the provided criteria, the text you provided is relevant to political content. It mentions "Pres Trump" as a political figure, discusses a policy issue related to wealth distribution and taxation. Therefore, it meets the criteria for RELEVANT to political content.

Question: I spoke on the floor of the House this morning to honor the life of longtime Syracuse Police Capt Richard Walsh\url{https://t.co/6CdNF5jcLc}

Answer: Based on the provided criteria, the text you provided is RELEVANT to political content. It mentions speaking on the floor of the House, honoring the life of a police captain, and includes a link to a source (\url{https://t.co/6CdNF5jcLc}). This text pertains to government activities, a political event (speech on the House floor), and mentions a political figure (Syracuse Police Capt Richard Walsh).

Now, is the following text relevant or irrelevant to political content?

[Paste a tweet here and remove the brackets]

\subsubsection{Task 2: Policy Frames \newline}
“Political content” refers to a text that pertains to politics or government policies at the local, national, or international level. This can include political figures, events, or issues, as well as text that uses political language or hashtags. 

I will ask you to classify a text as one of the frames defined below:

\begin{itemize}
    \item ECONOMY: The costs, benefits, or monetary/financial implications of the issue (to an individual, family, community, or to the economy as a whole).
    \item Capacity and resources: The lack of or availability of physical, geographical, spatial, human, and financial resources, or the capacity of existing systems and resources to implement or carry out policy goals.
    \item MORALITY: Any perspective—or policy objective or action (including proposed action)that is compelled by religious doctrine or interpretation, duty, honor, righteousness or any other sense of ethics or social responsibility.
    \item FAIRNESS AND EQUALITY: Equality or inequality with which laws, punishment, rewards, and resources are applied or distributed among individuals or groups. Also the balance between the rights or interests of one individual or group compared to another individual or group.
    \item POLICY PRESCRIPTION AND EVALUATION: Particular policies proposed for addressing an identified problem, and figuring out if certain policies will work, or if existing policies are effective.
    \item LAW AND ORDER, CRIME AND JUSTICE: Specific policies in practice and their enforcement, incentives, and implications. Includes stories about enforcement and interpretation of laws by individuals and law enforcement, breaking laws, loopholes, fines, sentencing and punishment. Increases or reductions in crime.
    \item SECURITY AND DEFENSE: Security, threats to security, and protection of one’s person, family, in-group, nation, etc. Generally an action or a call to action that can be taken to protect the welfare of a person, group, nation sometimes from a not yet manifested threat.
    \item HEALTH AND SAFETY: Health care access and effectiveness, illness, disease, sanitation, obesity, mental health effects, prevention of or perpetuation of gun violence, infrastructure and building safety.
    \item QUALITY OF LIFE: The effects of a policy on individuals’ wealth, mobility, access to resources, happiness, social structures, ease of day-to-day routines, quality of community life, etc.
    \item POLITICAL: Any political considerations surrounding an issue. Issue actions or efforts or stances that are political, such as partisan filibusters, lobbyist involvement, bipartisan efforts, deal-making and vote trading, appealing to one's base, mentions of political maneuvering. Explicit statements that a policy issue is good or bad for a particular political party.
    \item EXTERNAL REGULATION AND REPUTATION: The United States’ external relations with another nation; the external relations of one state with another; or relations between groups. This includes trade agreements and outcomes, comparisons of policy outcomes or desired policy outcomes.
    \item OTHER: Any topic that does not fit into the above categories.
\end{itemize}

The following tweets are some examples of these frames in the form of question and answer:

Question: HURRY!!! Today is the last day to sign up for health insurance that begins on Jan 1. Visit \url{https://t.co/rrKeGJOFBA} to \#GetCoveredNow. \#ACA \url{https://t.co/LCMQNHjCMN}

Answer: HEALTH AND SAFETY

Question: The \#CHOICEAct provides regulatory relief for community banks \& credit unions promoting more economic opportunity → \url{https://t.co/uOBmHKhrxkhttps://t.co/64WGHA1D2R}

Answer: ECONOMY

Question: The \#REINSAct signals our commitment to modeling reform that puts smart policy above tired politicking. \url{https://t.co/GpOcD1NZO7}

Answer: The text you provided falls under the frame of POLICY PRESCRIPTION AND EVALUATION, as it refers to the REINS Act and emphasizes the commitment to modeling reform and smart policy.

Question: Tonight it was my distinct privilege to speak on the Senate floor in support of my friend \& our Attorney General Jeff Sessions. \url{https://t.co/UoIYp1R3ES}

Answer: POLITICAL

Question: Thanks \@Astro\_Kate7 for speaking w/students at her Alma mater \@VHS\_Crusheds about her groundbreaking work on the International Space Station \url{https://t.co/UXnh8STwaN}

Answer: OTHER

Question: I always thought the best soldiers end up at Joint Base Lewis-McChord, but here's proof. Congrats to the 1st Special Forces Group (Airborne) sniper team! \@JBLM\_PAO \@TaskandPurpose \url{https://t.co/x8nX6HyYOQ}

Answer: SECURITY AND DEFENSE

Question: As I told \#SouthKorea leaders during my visit in Dec, US is committed to a strong alliance despite political turmoil \url{https://t.co/8orrFs8atv}

Answer: EXTERNAL REGULATION AND REPUTATION

Question: Proud to \#StandWithGavin and all transgender students. Every child deserves to go to school \& live as who they are free from discrimination. \url{https://t.co/4uqpuHzbCd}

Answer: Based on the provided text, the best frame that fits the text is "FAIRNESS AND EQUALITY." The text expresses support for transgender students and emphasizes the importance of ensuring that every child can go to school and live free from discrimination. This relates to the equality or inequality with which laws and resources are applied or distributed among individuals or groups. In this case, it pertains to the rights and interests of transgender students compared to others, highlighting the need for fairness and equal treatment.

Question: The prosecution of corruption by high ranking government officials, even years after the crimes were committed, is critical to...

Answer: Based on the provided text, the best frame that fits the text is "LAW AND ORDER, CRIME AND JUSTICE." The text discusses the prosecution of corruption by high-ranking government officials, which falls under specific policies in practice, their enforcement, incentives, and implications. It involves the enforcement and interpretation of laws, the accountability of government officials, and the pursuit of justice in addressing corruption.

Question: The Trump-Sessions "zero tolerance" family separation border policies are not required, right or moral. \url{https://t.co/aAFX8Q6eKT}

Answer: Based on the provided text, the frame that best fits is: MORALIT. The text expresses a moral perspective on the Trump-Sessions "zero tolerance" family separation border policies, stating that they are not required, right, or moral. It addresses the ethical dimension and social responsibility associated with these policies.

Question: Wisconsin is full of great role models and leaders. Congratulations to all of the outstanding women honored by the La Crosse YWCA, and thank you for making the coulee region a better place to live! \url{https://t.co/mj1HK4PwzI}

Answer: Based on the provided text, the frame that best fits is: QUALITY OF LIFE. The text highlights the positive impact of the outstanding women honored by the La Crosse YWCA in making the coulee region a better place to live. It emphasizes the effects of their contributions on the quality of life in the community.

Now, which of the above frames best fit the following text? Answer with only the option above that is most accurate and nothing else.

[Paste a tweet here and remove the brackets]

\subsection{Dataset 4: Content Moderation News Articles (2020-2021)}
\subsubsection{Task 1: Relevance \newline}

“Content moderation” refers to the practice of screening and monitoring content posted by users on social media sites to determine if the content should be published or not, based on specific rules and guidelines.

I will ask you to classify a text as relevant or irrelevant to the content moderation:

A: Text is RELEVANT if it includes: social media platforms’ content moderation rules and practices, censorship, governments’ regulation of online content moderation, and/or mild forms of content moderation like flagging, shadowbanning, or account suspension.

B: Text is IRRELEVANT if they do not refer to content moderation, as defined above. This would include, for example, a tweet by Trump that Twitter has labeled his tweet as “disputed”, or a tweet claiming that something is false.

The following texts are some examples of 'RELEVANT' or 'IRRELEVANT' texts to content moderation in the form of question and answer:

Question: TORONTO - Ontario Premier Doug Ford on Monday said the United States had blocked the delivery of nearly three million face masks at the American border over the weekend. Ford said restrictions on shipments at the U.S. border have left the province with just one more week's worth of personal protective equipment for health-care workers fighting the coronavirus outbreak in Ontario. In a statement today, he says Ontario is ramping up its own production of personal protective equipment, but most of those supplies are weeks away from being in the hands of front-line health workers. At least 451 health-care workers in Ontario have tested positive for COVID-19, representing about 10 per cent of all cases in the province. In all, Ontario reported 309 new COVID-19 cases today, including 13 new deaths. There have now been a total of 4,347 cases in the province, including 1,624 patients who have recovered and 132 deaths. Allies of the United States are complaining about its "Wild West" tactics in outbidding or blocking shipments to buyers who have already signed deals for medical equipment. Prime Minister Justin Trudeau sidestepped reporters' questions about the incident on Monday, saying his government was in productive talks with the United States and adding: "We expect

Answer: The provided text is 'IRRELEVANT' to content moderation. It discusses the United States blocking the delivery of face masks at the border, Ontario's shortage of personal protective equipment, COVID-19 cases and deaths, and complaints about the United States' tactics in outbidding or blocking shipments. It does not refer to social media platforms, content moderation rules, censorship, or any forms of online content moderation.

Question: A look at the first quarter of the year that was 2019. January 3: China's "Chang'e 4" is the first space probe to land on the far side of the moon. January 11: Macedonia is now North Macedonia, ending a row with Greece and paving the way for NATO membership and EU accession talks. The Greek parliament ratifies the historic name agreement on January 25.January 13: The mayor of the Polish city of Gdansk, Pawel Adamowicz, 53, is stabbed to death by a previously convicted bank robber during a fundraiser. January 15: Attackers claimed by the Somalia-based group al-Shabaab storm the upmarket Dusit hotel in the Kenyan capital Nairobi, killing more than 20. January 18: After four months of political deadlock in Sweden, Social Democratic leader Stefan Lofven wins a vote in parliament to form a government. January 18: At least 109 people are killed when a fuel pipeline explodes in the Mexican city of Tlahuelilpan. January 22: The EU Commission imposes a fine of 570 million euros on the credit card company Mastercard for artificially pushing up the cost of card payments. January 23: Juan Guaido, the head of Venezuela's opposition-dominated National Assembly, declares himself the country's interim president. January 24: Felix Tshisekedi is

Answer: The provided text is 'IRRELEVANT' to content moderation. It discusses various events that took place in the first quarter of 2019, including space exploration, political developments, criminal incidents, and international affairs. It does not refer to social media platforms, content moderation rules, censorship, or any forms of online content moderation.

Question: Bhubaneswar, Oct. 29 -- New Delhi: The Supreme Court has severely criticised the growing trend of police in different States summoning individuals from far corners of the country over social media postings. Hearing a petition concerning a Delhi resident Roshni Biswas who was reportedly summoned by Bengal Police for posting objectionable content on Facebook, the Bench of Justices DY Chandrachud and Indira Banerjee noted that police's power to issue summons under Section 41A of the Code of Criminal Procedure (CrPC) cannot be used to intimidate, threaten and harass. As per reports, the apex court's comment was prompted by Bengal Police issuing summons to the 29-year-old woman who, in a Facebook post, had criticised the Mamata government for non-enforcement of lockdown norms. The FIR which relies on FB links contains a statement that the posts implied the State administration was going soft on the violation of the lockdown at Rajabazar as the area is predominantly inhabited by a particular community and that the administration is complacent while dealing with lockdown violations caused by a certain segment of the community. Mahesh Jethmalani, learned senior counsel appearing on behalf of the petitioner submitted that the petitioner has stated on oath that she disclaims any association with

Answer: The provided text is 'RELEVANT' to content moderation. It discusses the Supreme Court's criticism of the trend of police summoning individuals from different states over social media postings. The text mentions the power of police to issue summons and emphasizes that it should not be used to intimidate or harass individuals. It also refers to a specific case involving objectionable content posted on Facebook and the implications of the posts regarding the state administration and lockdown violations. These aspects are related to content moderation rules, the actions of law enforcement, and the impact of social media postings. Therefore, the text falls within the scope of relevance to content moderation as defined in our previous classification.

Question: Facebook and Instagram took down video tributes to George Floyd posted by the Trump campaign over copyright complaints on Friday, following a similar decision by Twitter - broadening the latest skirmish over the policing of online messages from President Donald Trump and his allies. Facebook and its subsidiary Instagram removed posts by official Trump campaign accounts that included videos narrated by Trump discussing Floyd's death in Minneapolis. The narration is played over a series of photographs and videos that appear to have been taken during recent protests around the country over Floyd's killing, Politico reported. We received a copyright complaint from the creator under the Digital Millennium Copyright Act and have removed the post," Facebook Spokesperson Andy Stone told POLITICO in an email, adding, "Organizations that use original art shared on Instagram are expected to have the right to do so. The move by the companies follows a parallel action by Twitter, which on Thursday morning disabled the same video included in a pair of tweets by \@TeamTrump and \@TrumpWarRoom 2020 campaign accounts, also citing an unspecified complaint under the Digital Millennium Copyright Act. The company left up the tweets but placed an overlay on the videos that reads, This media has been

Answer: The provided text is 'RELEVANT' to content moderation. It discusses the actions taken by Facebook and Instagram to take down video tributes to George Floyd posted by the Trump campaign due to copyright complaints. The text mentions the removal of posts by official Trump campaign accounts on Facebook and Instagram, as well as the similar decision made by Twitter to disable the same video on its platform. These actions relate to content moderation rules and practices, specifically in terms of copyright infringement. Therefore, the text falls within the scope of relevance to content moderation as defined in our previous classification.

Now, is the following text relevant or irrelevant to content moderation?

[Paste a news article text here and remove the brackets]

\subsubsection{Task 2: Problem/Solution Frames \newline}

“Content moderation” refers to the practice of screening and monitoring content posted by users on social media sites to determine if the content should be published or not, based on specific rules and guidelines.

I will ask you to classify a text as describing content moderation as a problem, as a solution, or neither:

A: Text describes content moderation as a PROBLEM if they emphasize negative effects of it, such as restrictions to free speech, censorship, or the biases that can emerge from decisions regarding what users are allowed to post.

B: Text describes content moderation as a SOLUTION if they emphasize positive effects of it, such as protecting users from harmful content such as hate speech, misinformation, illegal adult content, or spam. 

C: Text describes content moderation as NEUTRAL if they do not emphasize negative or positive effects of content moderation. For example if they simply report on the content moderation activity of social media platforms without linking them to potential advantages or disadvantages for users or stakeholders.

The following texts are some examples of 'PROBLEM', 'SOLUTION' or 'NEUTRAL' texts about content moderation in the form of question and answer:

Question: Twitter removed a "misleading" tweet downplaying the efficacy of masks posted by a top coronavirus adviser to President Donald Trump, while U.S. cases surged before the Nov. 3 election, Trend reports citing Reuters. As the Trump administration fends off accusations that its mixed messaging on wearing masks hampered the fight against the coronavirus, Dr. Scott Atlas continued to minimize the importance of masks with a Twitter post on Saturday, saying, "Masks work? NO." Twitter Inc removed the tweet on Sunday, saying it violated its misleading information policy on COVID-19, which targets statements that have been confirmed to be false or misleading by subject-matter experts. The White House had no immediate comment on the decision. New infections have been rising fast in the United States, according to a Reuters analysis, with more than 69,400 reported on Friday, up from 46,000 a month ago. Total U.S. cases have surpassed 8 million. Trump, who was hospitalized with the disease for three nights in early October, has been criss-crossing the country in a surge of 11th-hour campaigning as he lags in many public opinion polls. His rallies draw thousands of supporters in close quarters, with many not wearing masks despite federal coronavirus guidelines. Despite data showing otherwise, Trump has said

Answer: The provided text can be classified as describing content moderation as a SOLUTION. It highlights how Twitter removed a tweet that downplayed the efficacy of masks, stating that it violated their policy on misleading information related to COVID-19. This demonstrates content moderation as a means to address and mitigate the spread of false or misleading information. By removing the tweet, Twitter aims to protect users from harmful content and ensure accurate information is shared regarding public health during the pandemic.

Question: OAKLAND, Calif. - Facebook has banned an extremist anti-government network loosely associated with the broader "boogaloo" movement, a slang term supporters use to refer to a second Civil War or a collapse of civilization. But the platform didn't try to name the group, underscoring the difficulty of grappling with an amorphous network linked to a string of domestic terror plots that appears to obfuscate its existence. Among other complications, its internet-savvy members tend to keep their distance from one another, frequently change their symbols and catch phrases and mask their intentions with sarcasm. The move by Facebook designates this group as a dangerous organization similar to the Islamic State group and white supremacists, both of which are already banned from its service. The social network is not banning all references to "boogaloo" and said it is only removing groups, accounts and pages when they have a "clear connection to violence or a credible threat to public safety." The loose movement is named after "Breakin' 2: Electric Boogaloo," a 1984 sequel to a movie about breakdancing. "Boogaloo" supporters have shown up at protests over COVID-19 lockdown orders, carrying rifles and wearing tactical gear over Hawaiian shirts - themselves a reference to "big luau," a

Answer: Based on the provided text, it can be classified as describing content moderation as a SOLUTION. The text highlights Facebook's action of banning an extremist anti-government network associated with the "boogaloo" movement, which is linked to domestic terror plots. Facebook's move is presented as designating the group as a dangerous organization, similar to the Islamic State group and white supremacists, and removing groups, accounts, and pages with a clear connection to violence or a credible threat to public safety. This portrays content moderation as a measure to protect public safety and prevent the spread of extremist content.

Question: Florida Governor Ron DeSantis announced this week that he would fine social media companies that ban political candidates. Every outlet from Fox News to MSNBC fired off missives about the bill. What got lost in the news coverage is that Silicon Valley deplatforms very few politicians, save shock-jocks like Donald Trump and Laura Loomer (if you want to call her a politician). The same cannot be said for sex workers. This month, Centro University released a study estimating that 46 percent of adult influencers reported losing access to Twitter or Instagram in the last year. The bans put a permanent dent in the stars’ income, with Centro estimating sex workers lose \$260 million a year due to social media bans. You won’t hear DeSantis, Fox News, Glenn Greenwald, or any other so-called free speech warriors decrying porn stars’ lost incomes, so let me break down how social media companies are screwing over porn stars (and not screwing them in a good way!). Silicon Valley titans have revoked my social media access multiple times. Take my recent Snapchat ban. The Santa Monica-based app barred me from posting on my public account, so I lost the means to communicate with fans who would

Answer: Based on the provided text, it can be classified as describing content moderation as a PROBLEM. The text criticizes social media companies for their content moderation practices, particularly regarding the ban on sex workers and adult influencers. It suggests that these bans have negative consequences, such as the loss of income for sex workers, and highlights the perceived lack of concern or attention from politicians and free speech advocates. The text emphasizes the negative effects of content moderation, including restrictions on speech and the impact on individuals' livelihoods, indicating that it views content moderation as a problem.

Question: TALLAHASSEE – Gov. Ron DeSantis' call for punishing social media sites that deplatformed former President Donald Trump narrowly cleared a Senate committee Monday and soon will be ready for a full vote in the Legislature. Sen. Jeff Brandes, R-St. Petersburg, was the lone Republican who argued against the proposal by fellow Republican Sen. Ray Rodrigues of Naples. Brandes labeled it a "big government bill." "This Senate is currently filled with small government Republicans who do believe that government shouldn't be in the lives of businesses," Brandes said. He added: "This is the exact opposite of the things that we stand for." But Rodrigues argued back that the measure doesn't defy free market principles. The bill (SB 7072) orders social media companies to publish standards with detailed definitions of when someone would be censored or blocked, and makes companies subject to as much as \$100,000 fines for deplatforming a Florida candidate. "I'm bringing you good policy supported by your constituents," Rodrigues said. The measure was approved 10-9 by the Appropriations Committee, its last stop before going to the Senate floor. A similar measure is ready for a full House vote. State and federal courts have generally taken a hands-off view involving regulating online platforms. Congress also has not

Answer: Based on the provided text, it can be classified as describing content moderation as a PROBLEM. The text highlights the debate and disagreement surrounding a proposed bill that aims to punish social media sites for deplatforming former President Donald Trump. Senator Jeff Brandes argues against the proposal, labeling it a "big government bill" and stating that it goes against the principles of small government Republicans. This indicates a negative view of content moderation, particularly in terms of government intervention and potential restrictions on businesses. While Senator Ray Rodrigues argues in favor of the bill, the emphasis in the text is on the opposition and criticism, suggesting a portrayal of content moderation as a problem.

Question: A scathing new report released by hedge fund Hindenburg Research claims that start-up Nikola is an 'intricate fraud' based on years of lies and fake products•Hindenburg claimed to have 'extensive evidence' that the company's proprietary technology was purchased from another company•The fund also accused Nikola's founder Trevor Milton of making countless false statements over the last decade and faking a product promotional video•When filming the Nikola One ad, the truck reportedly didn't have an engine•So the company reportedly rolled the prototype along a downhill stretch of a highway and filmed it as if it was being driven•In a tweet, Milton called the report a 'hit job' and asked the public to 'give me a few hours to put together responses to their lies'•Nikola's stock prices plummeted as much as 13 percent Thursday A scathing new report released by hedge fund Hindenburg Research claims that start-up Nikola is an 'intricate fraud' based on years of lies and fake products Hindenburg claimed to have 'extensive evidence' that the company's proprietary technology was purchased from another company The fund also accused Nikola's founder Trevor Milton of making countless false statements over the last decade and faking a product promotional video When filming the Nikola One ad, the

Answer: The provided text does not explicitly discuss content moderation or its effects. It focuses on a scathing report released by Hindenburg Research that accuses the startup company Nikola of fraud and making false statements. The text does not emphasize the positive or negative effects of content moderation, nor does it link the topic to potential advantages or disadvantages. Therefore, the text can be classified as NEUTRAL regarding the description of content moderation.

Question: Some of the toughest ads against Donald Trump are driven by lifelong Republicans unable to stomach the direction their party has taken. Washington: Rick Wilson apologises for running late for our phone interview: it's been a frantic morning for the veteran Republican ad-maker and his colleagues at the Lincoln Project. The anti-Trump group has just released its latest advertisement, slamming the US President for suggesting that the November 3 election may need to be delayed. In the half hour since the ad ??? titled We Will Vote ??? went live, it has already racked up more than 250,000 views online. That's nothing unusual for the operatives at the Lincoln Project, who have been pumping out attack ads at a prolific rate over recent months. "We push really fast all the time," Wilson says. "We drive ourselves and our team very hard because we think we are pursuing a worthwhile endeavour and we know it works." The group's co-founders include Steve Schmidt, who ran Republican nominee John McCain's 2008 campaign, and conservative lawyer George Conway, the husband of top Trump aide Kellyanne Conway. Having spent most of their adult lives working to get Republicans elected, they are now producing some of the toughest anti-Trump ads on

Answer: The provided text describes the Lincoln Project, an anti-Trump group that releases attack ads against the US President. While the text does not explicitly discuss content moderation, it focuses on the activities and efforts of the Lincoln Project in creating and disseminating ads. It does not emphasize the positive or negative effects of content moderation or link it to potential advantages or disadvantages. Therefore, the text can be classified as NEUTRAL regarding the description of content moderation.

Now, is the following text describing content moderation as a problem, as a solution, or neither?

[Paste a news article text here and remove the brackets]

\section{Supplementary Results} \label{appendix:full_results}


\begin{figure*}[h]
  \centering
  \includegraphics[width = 0.9\linewidth]{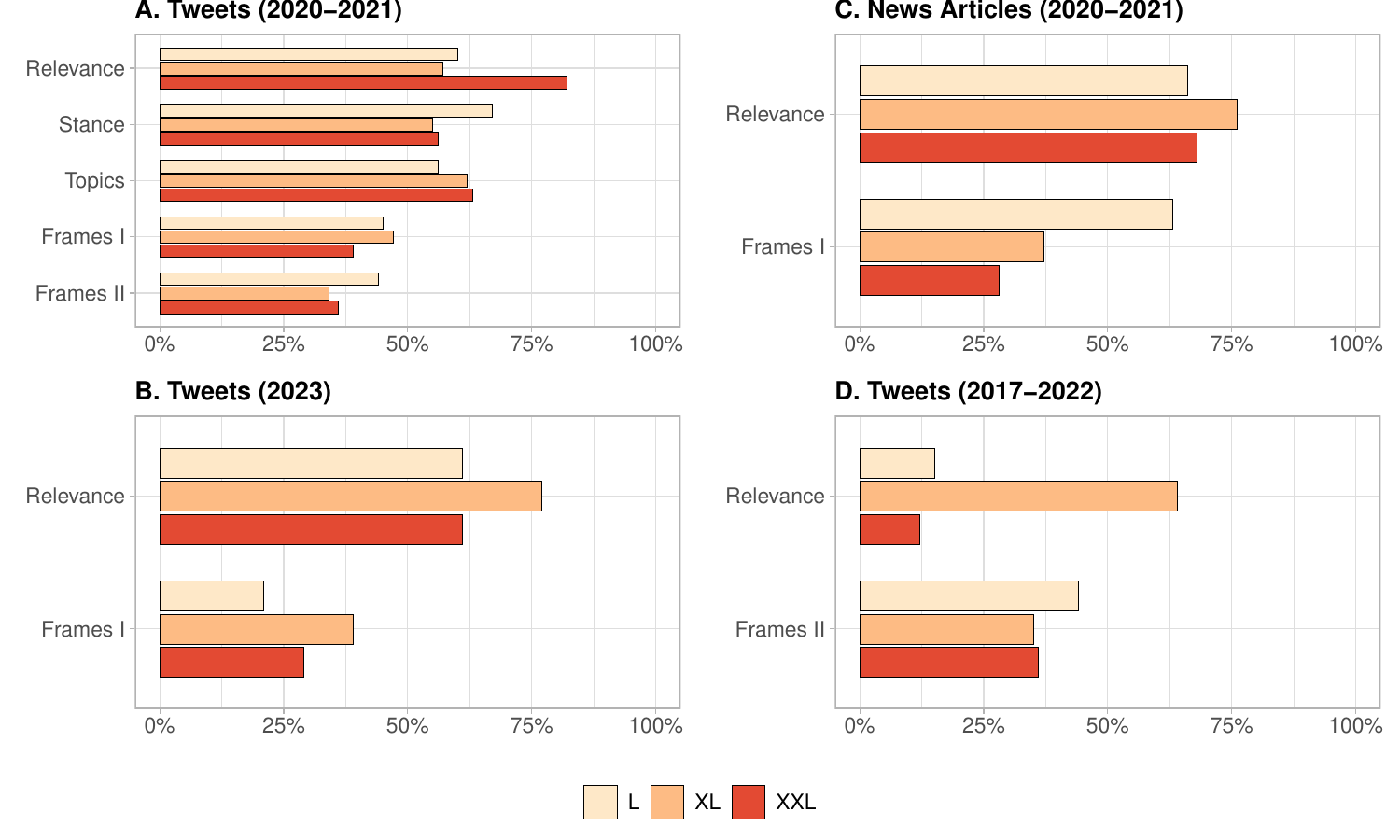}
  \caption{Comparing text annotation accuracy of FLAN-T5 (L), FLAN-T5 (XL), FLAN-T5 (XXL)}
  \label{flan_comp}
\end{figure*}

\begin{figure*}[h]
  \centering
  \includegraphics[width = 0.9\linewidth]{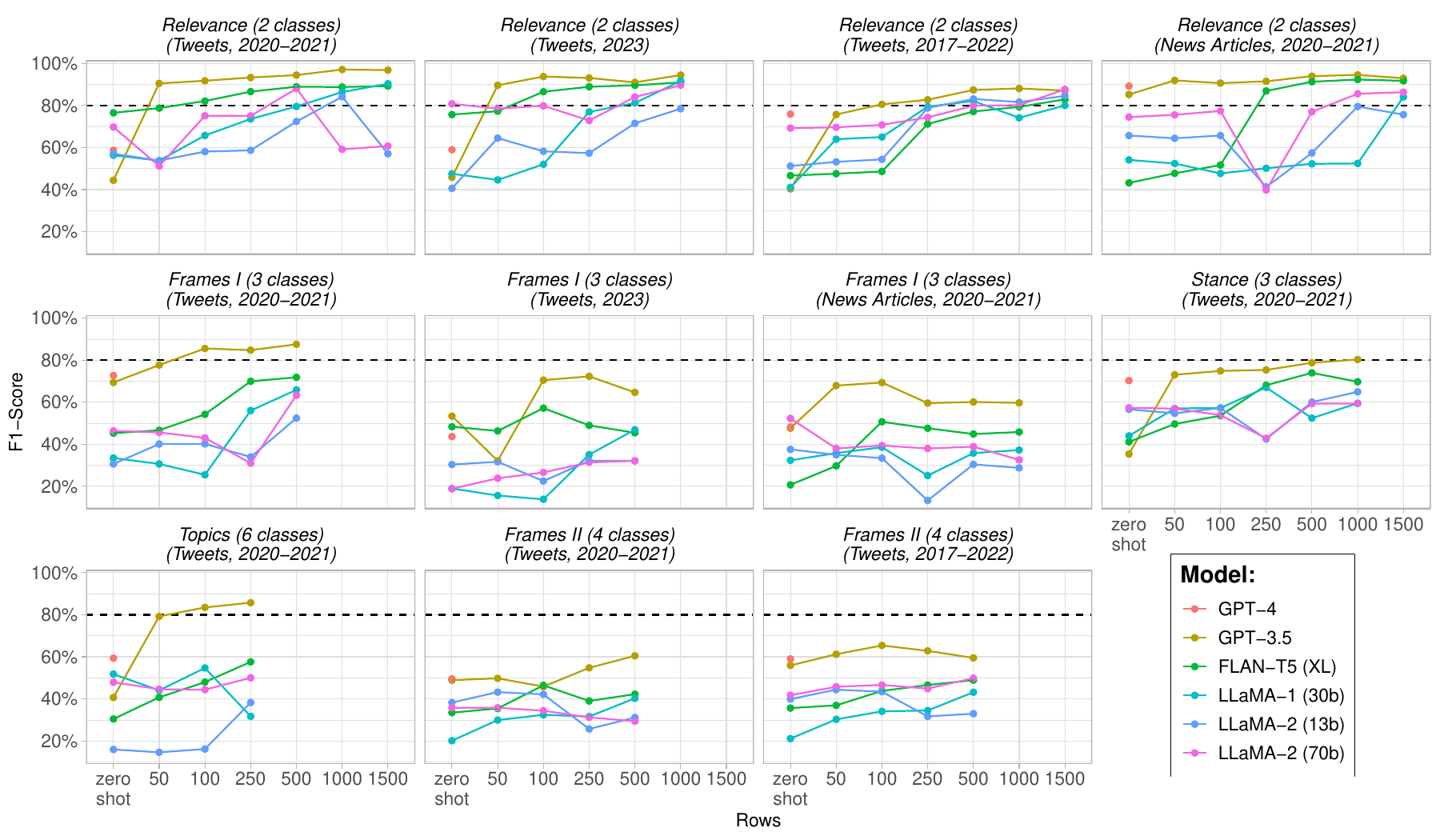}
  \caption{Performance (F1-Score) of GPT-3.5, LLaMA-1, LLaMA-2, and FLAN-T5 (XL), as a function of the
training data size for fine-tuning. The x-axis shows different sizes of training datasets, ranging from zero-shot (no
fine-tuning) to 50, 100, 250, 500, and 1,000 rows used for fine-tuning the models. The y-axis displays the accuracy
of the models in percentages.
Facets represent distinct tasks and/or datasets for evaluating the models}
\label{results_f1}
\end{figure*}

\begin{table}
    \center
    \setlength\tabcolsep{2.5pt}
    \caption{Best-performing model within each group (ChatGPT, LLaMA-1, FLAN) for each dataset and task. FLAN was run only zero-shot.} 
    \label{table:comparison}
    \begin{tabular}{lllll}
      \toprule
      Group & Shot & Version & Dataset & Task \\ 
      \midrule
      ChatGPT & few & temp 0.2 & News Articles (2020-2021) & Frames I \\ 
      ChatGPT & zero & temp 0.2 & News Articles (2020-2021) & Relevance \\ 
      ChatGPT & few & temp 0.2 & Tweets (2017-2022) & Frames II \\ 
      ChatGPT & few & temp 1 & Tweets (2017-2022) & Relevance \\ 
      ChatGPT & zero & temp 0.2 & Tweets (2020-2021) & Frames I \\ 
      ChatGPT & zero & temp 0.2 & Tweets (2020-2021) & Frames II \\ 
      ChatGPT & few & temp 1 & Tweets (2020-2021) & Frames II \\ 
      ChatGPT & zero & temp 1 & Tweets (2020-2021) & Relevance \\ 
      ChatGPT & zero & temp 0.2 & Tweets (2020-2021) & Stance \\ 
      ChatGPT & few & temp 0.2 & Tweets (2020-2021) & Topics \\ 
      ChatGPT & few & temp 0.2 & Tweets (2023) & Frames I \\ 
      ChatGPT & few & temp 1 & Tweets (2023) & Relevance \\ 
      FLAN & zero & L & News Articles (2020-2021) & Frames I \\ 
      FLAN & zero & XL & News Articles (2020-2021) & Relevance \\ 
      FLAN & zero & L & Tweets (2017-2022) & Frames II \\ 
      FLAN & zero & XL & Tweets (2017-2022) & Relevance \\ 
      FLAN & zero & XL & Tweets (2020-2021) & Frames I \\ 
      FLAN & zero & L & Tweets (2020-2021) & Frames II \\ 
      FLAN & zero & XXL & Tweets (2020-2021) & Relevance \\ 
      FLAN & zero & L & Tweets (2020-2021) & Stance \\ 
      FLAN & zero & XXL & Tweets (2020-2021) & Topics \\ 
      FLAN & zero & XL & Tweets (2023) & Frames I \\ 
      FLAN & zero & XL & Tweets (2023) & Relevance \\ 
      LLaMA-1 & zero & temp 0.2 & News Articles (2020-2021) & Frames I \\ 
      LLaMA-1 & zero & temp 0.2 & News Articles (2020-2021) & Relevance \\ 
      LLaMA-1 & few & temp 0.2 & Tweets (2017-2022) & Frames II \\ 
      LLaMA-1 & few & temp 0.2 & Tweets (2017-2022) & Relevance \\ 
      LLaMA-1 & few & temp 0.2 & Tweets (2020-2021) & Frames I \\ 
      LLaMA-1 & few & temp 0.2 & Tweets (2020-2021) & Frames II \\ 
      LLaMA-1 & few & temp 0.2 & Tweets (2020-2021) & Relevance \\ 
      LLaMA-1 & zero & temp 0.2 & Tweets (2020-2021) & Stance \\ 
      LLaMA-1 & few & temp 0.2 & Tweets (2020-2021) & Topics \\ 
      LLaMA-1 & zero & temp 0.2 & Tweets (2023) & Frames I \\ 
      LLaMA-1 & zero & temp 0.2 & Tweets (2023) & Relevance \\ 
  \bottomrule
\end{tabular}
\end{table}

\end{document}